\documentclass[10pt,twocolumn,letterpaper]{article}

\usepackage{iccv}
\usepackage{times}
\usepackage{epsfig}
\usepackage{graphicx}
\usepackage{amsmath}
\usepackage{amssymb}
\usepackage{graphbox}
\usepackage{booktabs}
\usepackage{multirow}
\usepackage{algorithm, algorithmic}
\usepackage{caption}
\usepackage{subcaption}

\usepackage{array}
\newcolumntype{@}{>{\global\let\currentrowstyle\relax}}
\newcolumntype{^}{>{\currentrowstyle}}
\newcommand{\rowstyle}[1]{\gdef\currentrowstyle{#1}%
  #1\ignorespaces
}

\usepackage[pagebackref=true,breaklinks=true,letterpaper=true,colorlinks,bookmarks=false]{hyperref}

\usepackage[capitalize]{cleveref}
\crefname{section}{Sec.}{Secs.}
\Crefname{section}{Section}{Sections}
\Crefname{table}{Table}{Tables}
\crefname{table}{Tab.}{Tabs.}
\iccvfinalcopy %

\usepackage{amsmath,amsfonts,bm}

\def\1{\bm{1}}

\def\vone{{\bm{1}}}

\def\vd{{\bm{d}}}
\def\ve{{\bm{e}}}
\def\vf{{\bm{f}}}
\def\vg{{\bm{g}}}

\def\vm{{\bm{m}}}
\def\vn{{\bm{n}}}

\def\vs{{\bm{s}}}

\def\vx{{\bm{x}}}
\def\vy{{\bm{y}}}

\DeclareMathAlphabet{\mathsfit}{\encodingdefault}{\sfdefault}{m}{sl}
\SetMathAlphabet{\mathsfit}{bold}{\encodingdefault}{\sfdefault}{bx}{n}

\def\sR{{\mathbb{R}}}

\newcommand{\E}{\mathbb{E}}

\def\iccvPaperID{7564} %

\ificcvfinal\fi

\begin{document}

\title{Towards Improved Input Masking for Convolutional Neural Networks}

\author{Sriram Balasubramanian\\
University of Maryland, College Park\\
{\tt\small sriramb@umd.edu}
\and
Soheil Feizi\\
University of Maryland, College Park\\
{\tt\small sfeizi@umd.edu}
}

\maketitle
\ificcvfinal\thispagestyle{empty}\fi

\begin{abstract}
     The ability to remove features from the input of machine learning models is very important to understand and interpret model predictions. However, this is non-trivial for vision models since masking out parts of the input image typically causes large distribution shifts. This is because the baseline color used for masking (typically grey or black) is out of distribution. Furthermore, the shape of the mask itself can contain unwanted signals which can be used by the model for its predictions. Recently, there has been some progress in mitigating this issue (called \textbf{missingness} bias) in image masking for vision transformers. In this work, we propose a new masking method for CNNs we call \textbf{layer masking} in which the missingness bias caused by masking is reduced to a large extent. Intuitively, layer masking applies a mask to intermediate activation maps so that the model only processes the unmasked input. We show that our method (i) is able to  eliminate or minimize the influence of the mask shape or color on the output of the model, and (ii) is much better than replacing the masked region by black or grey for input perturbation based interpretability techniques like LIME. Thus, layer masking is much less affected by missingness bias than other masking strategies. We also demonstrate how the shape of the mask may leak information about the class, thus affecting estimates of model reliance on class-relevant features derived from input masking. Furthermore, we discuss the role of data augmentation techniques for tackling this problem, and argue that they are not sufficient for preventing model reliance on mask shape. The code for this project is publicly available at \url{https://github.com/SriramB-98/layer_masking}.
\end{abstract}

\section{Introduction}
\label{sec:intro}

\begin{figure*}[ht!]
    \centering
    \includegraphics[width=\textwidth]{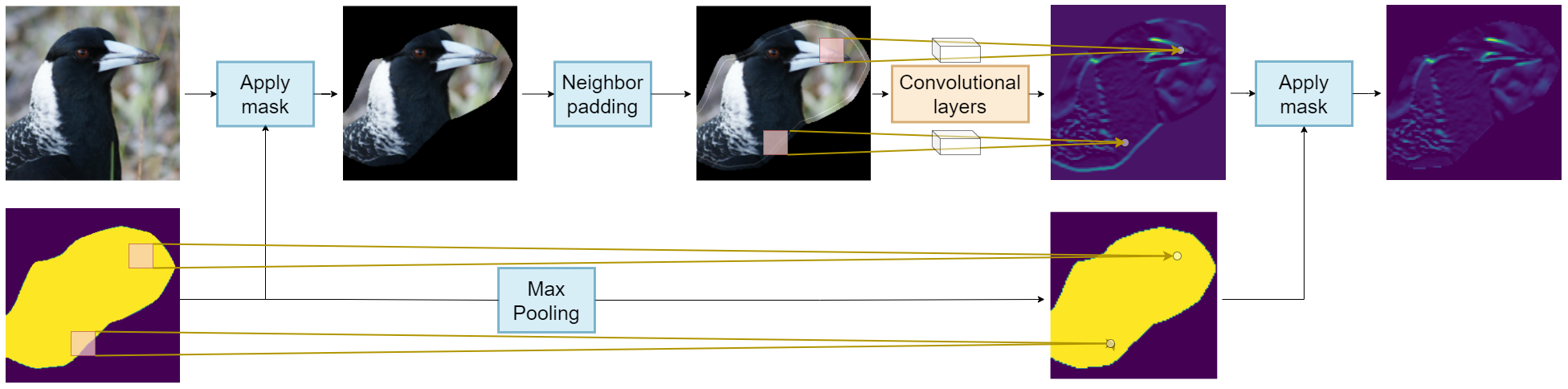}
    \caption{An outline of layer masking, our proposed method, for a convolutional layer. The image is first masked and then padded using neighbor padding. The convolutional layer then acts on the padded image, and a maxpool of the same kernel size and stride acts on the mask. These are then propagated forward through the CNN. The mask boundary is highlighted in the padded image for illustrative purposes.}
    \label{fig:master_seg}
\end{figure*}

While deep learning methods have become extremely successful in solving many computer vision tasks, they are generally opaque, and they do not admit easy debugging of errors. 
Many novel interpretability methods have been developed in recent years which attempt to analyze the rationale behind a model's predictions. 
In particular, it is natural to analyze the dependence of the model prediction on its input by perturbing parts of its input and observing corresponding changes in the output \cite{lime, shap, occlusion}. 
Common perturbations include adding Gaussian noise, Gaussian blurring, replacing with a baseline color, etc. %
However, many of these perturbation methods come with certain downsides. 
\emph{Partial perturbations}, like Gaussian noise or blurring, attempt to slightly corrupt parts of the image, while still preserving much of the information present in those parts.
While this has the advantage of not changing the input distribution drastically, we can only measure the local sensitivity of the model - if the model were to be robust to these perturbations, it would be locally insensitive to perturbations on certain parts of the image but it might still rely heavily on them for its prediction \cite{pmlr-v70-shrikumar17a, sundararajan2017axiomatic}. 
\emph{Full perturbation} methods remove the parts completely, and replace it with a baseline color like black or grey. In discrete domains like natural language, this is often the most popular method, as it is easy to remove words from the input \cite{pmlr-v130-mardaoui21a}. In images, however, this creates a large shift in input distribution, leading the model to perform poorly on such inputs \cite{sturmfels2020visualizing, sundar_note}.
For example, if we randomly mask out 16 $\times$ 16 sized patches from the image, ResNets are more likely to predict that the image is a maze or crossword \cite{jain2022missingness}

In recent work \cite{naseer2021intriguing, vit_adv_rob, vit_patch_rob}, it has been observed that vision transformers \cite{dosovitskiy2021an} are highly robust to many kinds of large magnitude input perturbations like occlusions and domain shifts, maintaining upto 60\% accuracy on ImageNet even if 80\% of the input is randomly blacked out. Jain et al \cite{jain2022missingness} argue that this property can make interpretability methods based on full perturbation especially effective for transformers. They further propose to simply drop tokens corresponding to masked out input parts instead of blacking out or greying out image portions, just like dropping BPE tokens in a transformer-based language model. This would make the transformer model completely insensitive to choice of baseline color and the shape of the mask.

Motivated by the same intuition, we devise a new masking technique for CNNs which mitigates the drawbacks of full perturbation to a large extent,  which we call \textbf{layer masking}. Layer masking (as depicted in \cref{fig:master_seg}) works by running the CNN only on the unmasked portion of the image, thus avoiding any large distribution shift. 
This is done by carefully masking and padding the input of each layer to make the model focus only on the unmasked input regions. 
Using this technique, we are able to randomly remove upto 50 \% of the input to a ResNet-50 (in the form of $16 \times 16$ sized patches) while maintaining the top-1 accuracy on ImageNet over 70\%. We are also able to mask out objects from images precisely without leaking any information about those objects via the shape of the mask.  
In addition, layer masking operates at the pixel level and is thus much more flexible than token dropping for vision transformers which only acts on a patch level. We also find that LIME \cite{lime} scores obtained using our masking method are more aligned with the most salient features of the image as compared to simply blacking or greying out the masked portion.

\section{Related Work}
\label{sec:rel_work}

Many interpretability methods designed for computer vision \cite{lime, shap, occlusion, fong2017interpretable} and prior work which attempts to quantify the reliance of the model on various features \cite{moayeri2022hard, xiao2020noise, moayeri2022comprehensive, wang2019balanced} analyze model predictions by implicitly relying on the ability to remove features from the input. Frequently, the removed input features are either masked out and replaced by a ``neutral" baseline color like black or grey, or perturbed partially by blurring or adding Gaussian noise. 
It has been shown in \cite{sundar_note, sturmfels2020visualizing} that many of these baseline colors are not really neutral, and interpretability methods which rely on this notion can often be quite sensitive to choice of color and the shape of the mask. While partial perturbation methods like adding Gaussian noise may not have the same issues, they can be misleading when the model is insensitive to such perturbations and its output doesn't change significantly  \cite{pmlr-v70-shrikumar17a, sundararajan2017axiomatic}.
To solve the problem of distribution shift created due to the masking patterns, \cite{hooker_roar} suggested retraining the model with the input perturbed according to the masks utilized in the interpretability methods. While this indeed solves the problem, there are two downsides: (1) the retrained model is not the original model and only a surrogate, and thus may not be very useful for interpretability, (2) retraining the model on masked input data is expensive and may even be infeasible if there is high variation in the shape of the masks to be used at inference time. We could also inpaint the masked region using a deep generative model \cite{chang2018explaining} to produce natural images, but this requires training a generative model which can be expensive. It also may leak hidden information - for example, if a dog's snout was masked out using this method, the generative model may regenerate the dog's snout again if that is the most likely completion.

 Several recent works \cite{naseer2021intriguing, vit_adv_rob, vit_patch_rob, jain2022missingness} have shown that vision transformers can be very robust (especially compared to CNNs) to many kinds of perturbations including occlusions, patch permutation, adversarial perturbations, distribution shift, etc.  According to \cite{jain2022missingness}, interpretability methods for vision transformers are less affected by masking patterns and baseline colors. Additionally, it is possible to drop the required patch tokens instead of replacing them with a baseline color. We devise a similar method for CNNs in this work and make progress towards bridging the interpretability gap between CNNs and transformers.%

While we focus on potential contributions to model interpretability and explainability, image masking is important in many other contexts. Image masking can be used to mitigate the reliance of the model on spurious correlations \cite{asgari2022masktune} by masking out spurious features while training the model. Several other works \cite{wang2017residual, zheng2017learning, sharma2015action} have also proposed using masks to eliminate irrelevant data using training. Defenses against patch attacks \cite{levine_patch, cert_patch, liu2022segment, yatsura2022certified} also utilize random masking to defend against adversaries. We believe that better masking techniques may be useful in these cases too.

We also note that methods similar to layer masking like partial convolution \cite{liu2018image} have been proposed for image inpainting. As we show in the supplementary, they do not work as well in the context of model interpretability because of the added constraint we face that we cannot retrain the model on our masking method.

\section{Proposed Method: Layer Masking }
\label{sec:method}

\subsection{Motivation}

We design a novel feature masking technique for CNNs which we call \emph{layer masking}. Given a model, an input image, and a mask for the input image, we aim to compute the model output such that (1) it doesn't depend on the masked out portion of the input and (2) it only depends on the unmasked portion of the input, and not the mask itself.

Modern CNNs primarily consist of convolutional layers, along with other layers like batch normalization, max pooling, average pooling, ReLU activations, etc. We can categorize these layers according to the size of their receptive fields. Layers with small receptive fields include convolutional layers, max-pooling and average pooling layers with kernel size much smaller than the size of the image. Fully connected layers, on the other hand, have a large receptive fields as each output depends on all the inputs. Layers with a small receptive field are in general more interpretable because they have fewer parameters and are implicitly hierarchical: for example, a stack of convolutional layers with a small kernel size processes  local information first and then progressively expands its receptive field to encompass the whole image.
We exploit this structure by devising an algorithm which carefully masks the input and output at \textit{each layer with small receptive field} such that information loss and artifacts created by the masking procedure at each step is minimal. We propagate both the input and the mask at each layer so as to simulate running a CNN on an irregularly shaped input corresponding to the unmasked input features, rather than substituting the masked inputs with a baseline color. We are careful, however, to \textbf{not} propagate forward any information in the masked out input regions.

Let the input to a convolutional layer with small receptive field  be $\vx \in \sR^{c \times n \times n}$ with output $\vy \in \sR^{c' \times n' \times n'}$ and binary input mask $\vm \in \{0,1\}^{n \times n}$ ($\vm[u,v]=1$ implies that cell $(u, v)$ is unmasked, else it is masked out).
Each element of the output of this layer with kernel size $k \times k$ depends on at most $k^2$ input values. These input values may either be all masked, all unmasked or partially masked and unmasked (when the convolution is over the mask edge), depending on the values of $\vm$ over the receptive field.

\begin{figure}
    \centering
    \includegraphics[width=0.48\textwidth]{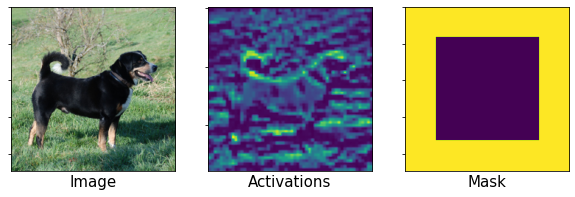}
    \includegraphics[width=0.48\textwidth]{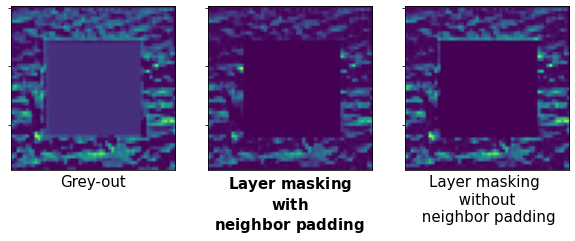}
    \caption{\textbf{Top row:} A dog image, sample activations after the first residual block, and mask to be applied to the image. \textbf{Bottom row:} The same activations when using grey-out masking, layer masking with neighbor padding (our method), and layer masking without neighbor padding (using zero padding). Neighbor padding helps in eliminating undesirable edge artifacts encountered in zero padding and greying out. Layer masking completely zeros out the masked out region unlike greying out which has non-zero values after a few layers}
    \label{fig:example_acts}
\end{figure}

It is clear that our masking procedure should propagate forward the outputs which only depend on the unmasked input, and  discard those outputs which depend only on the masked portion. However, it is not immediately obvious how to handle the outputs from the convolutions over the mask edge. The challenge here is that edge convolutions contain valuable information about the edges, and if we discard them at each layer, the unmasked portion of the image can quickly vanish to zero. Thus, we choose to propagate forward the edge convolutions. However, there is the danger of them distorting the natural distribution of the layer activations, as the output unavoidably depends on the masked out region which is filled with zeros. For example, in the third figure (bottom row) of \cref{fig:example_acts}, we see a slice of the activations obtained after applying the 1st residual block of ResNet-50 on the image with the central square region masked out at every layer, but including all the edge convolutions in the output. We see that the convolutions at the top edge of the mask result in a brighter top edge which indicates high activations. This is undesirable since this is an artifact created due to the masking method. We hypothesize that this is because the abrupt transition between the unmasked input and zeros trigger the filters sensitive to edges, thus creating a large activation.

\begin{figure}
    \centering
    \includegraphics[width=0.45\textwidth]{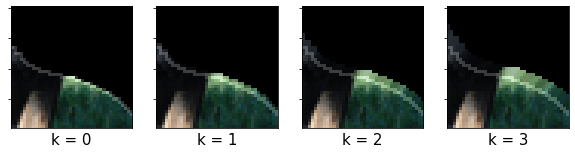}
    \caption{A visual depiction of neighbor padding on a part of the image as $k$ increases. The grey line is the mask edge (added in for illustrative purposes), the cells near the edge are filled progressively with the average of their neighbors' values }
    \label{fig:neighbor_pad}
\end{figure}

To mitigate this issue, rather than just fill the masked out portion with zeros, we pad the unmasked portion using a variant of replication padding we call \textbf{neighbor padding}. Specifically, we iteratively assign the masked input cells adjacent to the mask edge with the average value of its immediate non-zero neighbors. This process is continued till the width of the padding is at least $k$, the kernel size of the layer. In \cref{fig:neighbor_pad}, we see that after the cells near the edge are progressively filled using the values of its neighbors, the resultant image looks very natural and there is no sharp discontinuity near the edge. In an ablation study (see  supplementary), we find that this works much better than padding with zeros.

\begin{algorithm}[h]
\caption{Neighbor padding algorithm ($\text{Pad}_k(\vx, \vm)$)}
\label{algo:neighbor_pad}
\begin{algorithmic}
   \STATE {\bfseries Input:} Input to be padded $\vx$, Mask $\vm$, padding width $k$
   \STATE {\bfseries Output:} Padded input $\vx'$
   \STATE Initialize $\vx' \gets \vx \odot \vm$, \quad$\epsilon \gets 10^{-8}$
   \STATE Initialize $\vf \gets \vone_{3 \times 3}$,  a $3 \times 3$ filter filled with ones
   \FOR{$i=1$ {\bfseries to} $k$}
        \STATE $\vn \gets \vx' * \vf$ //  Numerator of the neighbor average
        \STATE $\vd \gets \vm * \vf$ // Denominator of the neighbor average
        \STATE $\ve \gets (1- \vm) \odot \vn / (\vd + \epsilon)$ // Fill masked inputs
       \STATE $\vx' \gets \vx' + \ve$
       \STATE $\vm \gets \min(\vone, \vm + \vd)$ // Update masks
   \ENDFOR
    
\end{algorithmic}
\end{algorithm}

We also have to propagate the masks forward, such that for the output of any layer, the corresponding mask is of the same shape as the output and indicates which output values need to be masked out by the following layers. Since  edge convolutions are not discarded at any step, the propagated mask must contain 1 for all output cells which depends on the unmasked portion of the input, and 0 everywhere else.

\subsection{Formal Description}
We now describe our method more formally. Suppose we are given a CNN $f$ which is structured like a directed acyclic graph. Each node of the DAG represents a layer or operation which acts on the outputs of the nodes with which it has an incoming edge. We replace each layer with its masking version (subscripted with $m$) which acts on an input-mask pair. 
Let $g_k$ be a layer with receptive field of size $k$. Then, we define its masking version: 

\vspace{-20pt} 
$$g_{k, m}(\vx, \vm) = (g_k(\text{Pad}_k(\vx, \vm)),  \text{MaxPool}_k(\vm)) )$$
\vspace{-14pt} 

 In this equation, $g_k$ could be any convolutional or pooling layer with kernel size $k$ and some stride $s$. $\text{MaxPool}_k$ is a max pooling layer with the same  kernel size and stride as $g_k$. $\text{Pad}_k(\vx, \vm)$ is a function which neighbor pads $\vx \odot \vm$  with padding width $k$ (described in \cref{algo:neighbor_pad}). Here, $\odot$ is the Hadamard product (with suitable broadcasting), and $*$ is convolution with zero padding. The max pool layer ensures that the output masks contains a 1 for all convolutions where even a single input was unmasked.

Layers which act independently on each element (like ReLU and BatchNorm) can  be considered to be a special case of the above with $k=1$.  In this case, the above equation is greatly simplified and becomes: 
\vspace{-10pt}

$$g_{m}(\vx, \vm) = (g(\vx \odot \vm),  \vm)$$

\vspace{-3pt}
In models which use residual connections,  two input - mask pairs can be added together as:
\vspace{-8pt}
$$( \vx_1, \vm_1) + ( \vx_2, \vm_2) = ( (\vx_1 + \vx_2) \odot (\vm_1 \odot \vm_2), \vm_1 \odot \vm_2)$$

\vspace{-5pt}
We lose some information here by taking the Hadamard product of the masks, but this is negligible in practice.

The penultimate layer is generally a global average pooling layer which averages over the height and width of the activation maps and return a single number. If $h$ is a global average pooling layer, then we define its masking version:
\vspace{-8pt}
 $$h_{m}(\vx, \vm) = h(\vx \odot \vm)/h(\vm)$$
The layer's output is rescaled by the mean value of the mask, which ensures that the output's magnitude is comparable to when there is no masking. Such layers may also be utilized for recalibrating channel-wise features by multiplying the activation maps with the output of the average pooling layer (like in Squeeze Excitation blocks \cite{hu2018squeeze}). Layers after the penultimate global average pooling layer act on the input as normal.

We can now create a new model $f_m$ which has the same DAG structure of the original model $f$, except each layer $g^i$ or $h^i$ has been replaced with the corresponding masking version $g^i_m$ or $h^i_m$. $f_m$ acts on an image - mask pair and produces an output which depends only on the unmasked portion of the image.

\section{Experiments}
\label{sec:exps}

We examine the performance of layer masking compared to baselines on three dimensions: (1) the robustness of 
models as increasingly larger portions of the image are masked out; (2) the effect of mask shape on model prediction when the shape of the mask reveals information about the hidden object; and (3) the effect of masking method on LIME. We also examine the role of data augmentation during pretraining on missingness bias of different masking methods. We use  \textbf{grey-out} (replace the masked out input with a grey color equal to the ImageNet mean) and \textbf{black-out} (replace the masked out input with a black color) as baselines, and focus on ResNet-50 \cite{DBLP:journals/corr/HeZRS15} in this section (results on other CNN architectures in the supplementary). We evaluate all models on the ImageNet dataset, with segmentation masks from Pixel ImageNet \cite{zhang2020interactive} and saliency maps from Salient ImageNet \cite{singla2022salient}.

\subsection{Segment masking experiments}
\label{sec:seg_abl}

\begin{table*}[ht]
    \centering
    \begin{tabular}{@p{24mm}|^p{10mm}^p{13mm}^p{13mm}|^p{10mm}^p{13mm}^p{13mm}|^p{10mm}^p{13mm}^p{13mm}}
     \hline
                  & \multicolumn{3}{c|}{Quickshift segments} &  \multicolumn{3}{c|}{$16 \times 16$ patches} &  \multicolumn{3}{c}{SLIC superpixels}  \\
                  &  Random & Most sal. first & Least sal. first &  Random & Most sal. first & Least sal. first &  Random & Most sal. first & Least sal. first\\
                  \hline
     \hspace{-5pt}Accuracy &&&&&&&&&\\
        Blackout &  0.395 &  0.124 &  0.767 &  0.181 &  0.340 &  0.582 &  0.347 &  0.200 &  0.657 \\
        Greyout &  0.463 &  0.137 &  0.787 &  0.240 &  0.374 &  0.621 &  0.418 &  0.234 &  0.702 \\
        \rowstyle{\bfseries} Layer masking &  0.551 &  0.159 &  0.806 &  0.577 &  0.484 &  0.703 &  0.542 &  0.294 &  0.756 \\
         \hline
        \hspace{-5pt}Class entropy &&&&&&&&&\\
       Blackout &  4.988 &  2.224 &  5.824 &  2.574 &  4.570 &  5.406 &  4.815 &  3.976 &  5.661 \\
Greyout &  5.327 &  2.362 &  5.875 &  3.289 &  4.807 &  5.642 &  5.022 &  4.229 &  5.808 \\
 \rowstyle{\bfseries} Layer masking &  5.698 &  2.572 &  5.892 &  5.651 &  5.782 &  5.879 &  5.607 &  4.763 &  5.876 \\
         \hline
    \end{tabular}
    \caption{AUC of the accuracy \textbf{(top)} and class entropy of the predictions \textbf{(bottom)} vs fraction of segments masked out }
    \label{tab:seg_ablate_auc}
\end{table*}

\begin{figure}[ht]
    \centering
    \includegraphics[width=0.49\textwidth]{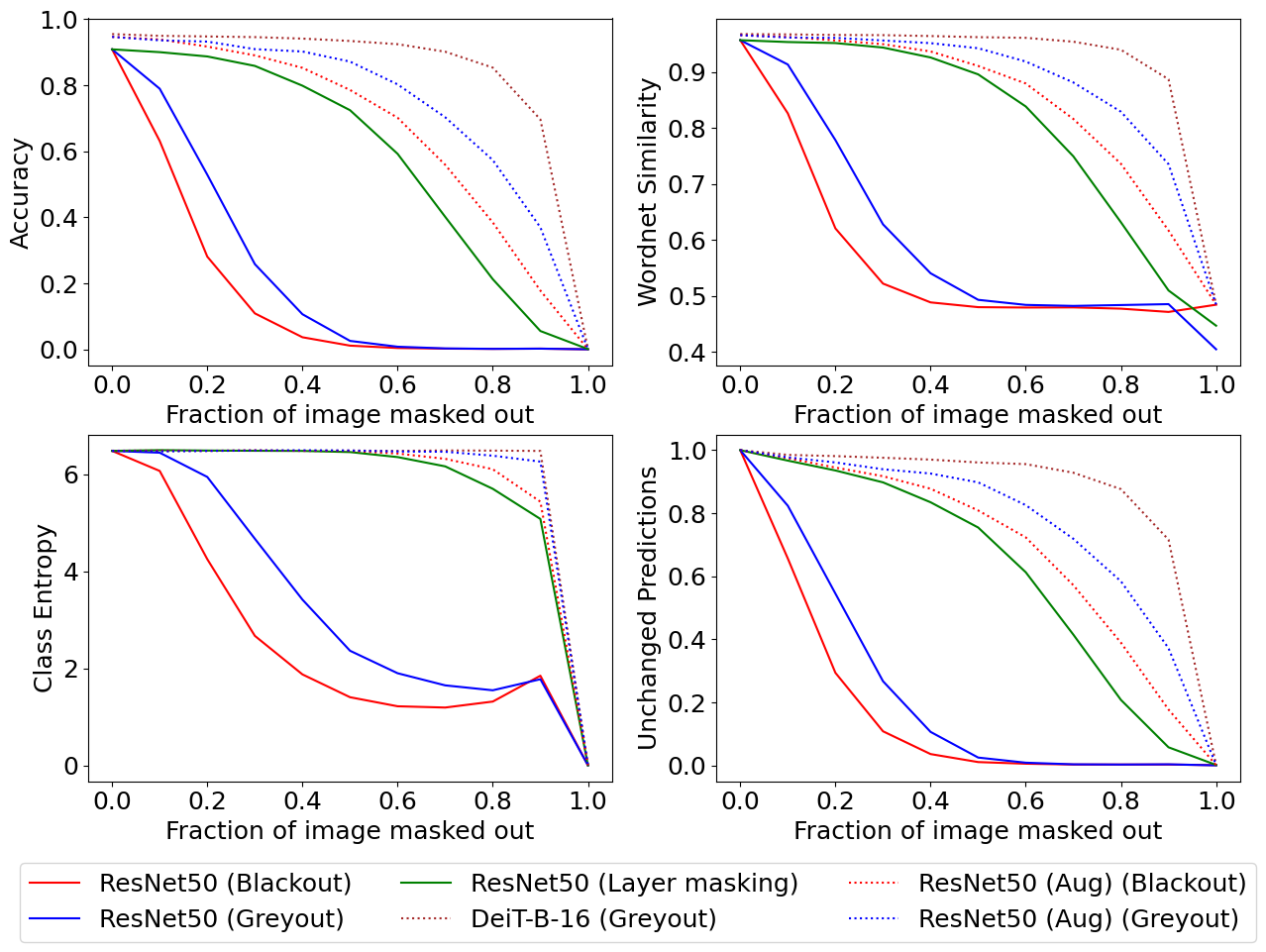}
    \caption{Metrics plotted as a function of fraction of $16 \times 16$ patches masked out in a random order using a given masking method and model. ResNet50 (Aug) refers to ResNet50 pretrained with grey missingness augmentations}
    \label{fig:all_metrics}
\end{figure}

To quantify the effect of feature masking methods on the model predictions, we study the behavior of the model when varying parts of the input  are masked out. We first segment the image using a segmentation algorithm. Then, we analyze how the model output changes when more and more segments are masked out using a given masking technique.  We characterize the model behavior using 4 metrics: accuracy, class entropy (defined as entropy of $p_f(y) = \E_{\vx \in D} [ 1[f(\vx) = y] ]$), WordNet similarity \cite{miller-1994-wordnet} between predictions and true labels, and fraction of unchanged predictions. The WordNet similarity measures how similar the model predictions are to the true labels on a scale from 0 to 1 in place of a binary hit / miss. The fraction of unchanged predictions is a measure of the number of predictions changed by masking out parts of the input image. The class entropy indicates if the predictions are skewed towards a particular class or if they are equally distributed.

We also use the following segmentation algorithms to extract the segments from the image (1) \textbf{Square patches}: Segment the $224 \times 224$ image into smaller $16 \times 16$ square patches , (2) \textbf{SLIC} \cite{6205760}, and (3) \textbf{Quickshift} \cite{quickshift}. We tune hyperparameters for these algorithms such that they divide the image into approximately the same number of segments. 

As in previous work \cite{naseer2021intriguing, jain2022missingness}, we mask out these segments in three orders using their saliency scores: (1) \textbf{Randomly}, (2) \textbf{Most salient first}, (3)\textbf{ Least salient first}. 
To compute the saliency scores, we select saliency maps from Salient ImageNet \cite{singla2022salient}. Each saliency map $m \in \sR^{d \times d}$ is a pixel level saliency attribution array where $0 \leq m[i, j] \leq 1$  denotes the importance of the $(i, j)$th pixel to predicting the ImageNet class - the higher the value, the more salient the pixel. We then compute saliency scores for each segment by adding the saliency values for all pixels in that segment.

We then evaluate the metrics listed above, and plot them as a function of the fraction of segments masked out. The plots obtained by removing $16 \times 16$ sized patches from the images in a random order can be found in \cref{fig:all_metrics}, while the rest can be found in the supplementary. The area-under-curve (AUC) of the accuracy and class entropy vs fraction of the segments masked for different baselines and order os masking can be found in \cref{tab:seg_ablate_auc}.

Ideally, the masking technique should be such that the model ignores the masked out region completely. Thus, any performance drop of the model due to distribution shift should be minimal. This distribution shift can come about due to the unnatural baseline color and/or the shape of the mask. The less rapidly the metrics degrade, the more robust the model and the masking technique. 

\cref{fig:all_metrics} and \cref{tab:seg_ablate_auc} indicate that ResNet-50 is much more robust over all metrics when the segments are removed using layer masking as opposed to black-out/grey-out. Greying out is also found to be better than blacking out, as expected. This difference in robustness persists across various segmentation and order of segment removal, and is the highest when $16\times 16$ sized patches are removed. Surprisingly, removing random $16 \times 16$ patches degrades accuracy and class entropy more rapidly as compared to masking out the most salient regions first. This is because scattered blackout patches strongly resemble a maze or crossword pattern, while most salient regions are contiguous and do not resemble a maze as much. Thus, the model is confused by the shape of the mask and predicts incorrectly. This shows that the shape of the mask may also be a factor which contributes to missingness bias. We discuss this issue in \cref{sec:shape_bias}.

Consistent with \cite{naseer2021intriguing}, we find that DeiTs are still more robust than ResNets - even when utilizing layer masking. Also, when ResNet-50 is pre-trained with data augmentations like RandAugment containing grey missingness approximations as in \cite{wightman2021resnet}, performance drop due to distribution shifts is reduced dramatically. DeiTs also benefit from these data augmentations which makes them more robust than plain ViTs. However, we argue that data augmentation is only a partial solution for the problem of missingness bias. We discuss this in more detail in \cref{sec:data_aug_disc}

\subsection{Effect of mask shape on model prediction}
\label{sec:shape_bias}

\begin{figure}[ht]
    \centering
    \includegraphics[width=0.49\textwidth]{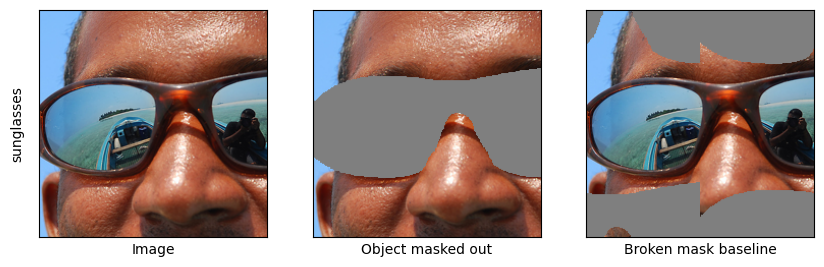}
    \includegraphics[width=0.49\textwidth]{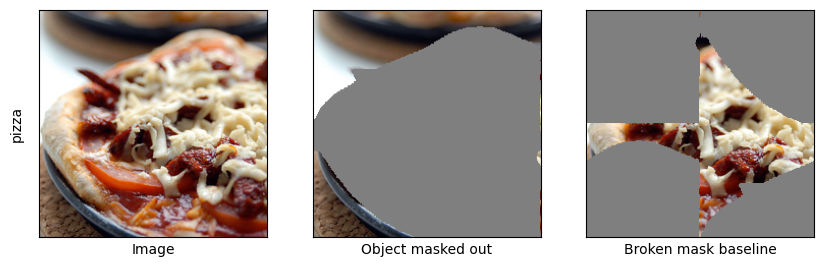}
    \caption{(\textbf{Left}) Two sample images from ImageNet, (\textbf{Middle}) with their relevant objects masked out, and (\textbf{Right}) the images with masks broken into four pieces and switched around }
    \label{fig:example_shape_mask}
\end{figure}

\begin{figure*}[ht]
    \centering
    \includegraphics[width=\textwidth]{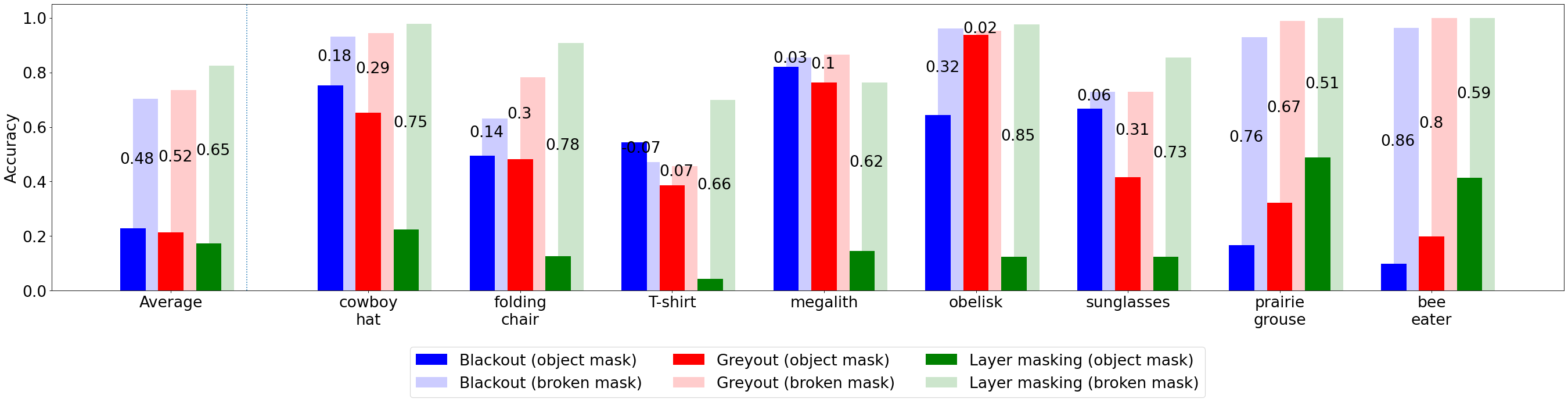}

    \caption{Accuracy of ResNet-50 over some salient classes when the relevant object(s) is masked out with various masking techniques (in dark colors) as compared to when the mask is broken into four (in light colors). `Average' denotes average accuracy over all classes. The accuracy difference between broken mask and object mask cases is printed on the bars.}
    \label{fig:shape_bias_barplot}
\end{figure*}

We now examine the extent to which the model output relies on the shape of the mask itself when only the relevant objects are masked out from the input. For example, the sunglasses-shaped mask in \cref{fig:example_shape_mask} may signal to the model that the object being masked out is sunglasses. Using a different mask shape can potentially decrease or remove this shape dependence but there is a possibility that this approach might unintentionally conceal areas of the input image that were meant to remain visible.  In many cases, this is not a significant problem. However, when two significant objects are in close proximity, we may need to mask out one object while leaving the other unmasked. For example, we might require the sunglasses mask to \textit{not} cover the nose or ears to evaluate the model's dependence on these features. We also do not know if the chosen mask shape (say a rectangle) is associated with any class (like `box' or `crate'). Thus, we require a masking technique which can precisely remove specific regions of the input but not leak any information about the masked input. Layer masking seems like a promising candidate, as it ensures that the masked regions are not processed by the model at all. 

To quantify this effect, let us analyze the distribution shift introduced by masking in detail. There are primarily three components to it: (a) the removal of salient image content from the masked out region, (b) introduction of a new baseline color in place of the original content, and (c) the  shape of the mask. All three can contribute to drop in accuracy after masking. However, our goal is to evaluate the effect of (c) on the accuracy. Suppose that we now compute the accuracy of the model on correctly classified images after masking the relevant objects using segmentation masks (second column, \cref{fig:example_shape_mask}). All masking methods apply the same mask and remove the content completely, so we already control for (a). 
However, different masking methods can differ w.r.t (b), which means that the differences in accuracy drop after masking cannot be attributed to (c) alone. We control for this by also computing accuracy after breaking up the mask into four pieces (of size $112 \times 112$) and sending them to the opposite corners (third column, \cref{fig:example_shape_mask}). We expect there to be little useful signal in the shape of the broken mask and any accuracy drop should arise from (a) and (b) only. Effect of (a) is the same across masking methods, so (a) is controlled. The effect of (b) in both cases is very similar as the area of the broken mask is the same as the original mask's area, so the difference in accuracy between the two cases (broken mask vs. object mask) should capture the effect of (c) on the model predictions. We expect (c) to have a positive impact on the accuracy (and thus negative impact on the extent of accuracy drop), as the shape of the object mask is a useful signal which can help with class prediction. Therefore, the larger the accuracy difference between the object mask and broken mask cases, the lower the effect of (c) on the model prediction after masking.

We carry out this experiment using images and segmentation masks from Pixel ImageNet  \cite{zhang2020interactive}. We observe in \cref{fig:shape_bias_barplot} that layer masking has the lowest average accuracy on the object masked images, but the highest accuracy on the broken mask baseline which indicates that it has the lowest reliance on mask shape. We then pick a few classes over which the accuracy difference is either much lower or much higher for layer masking as compared to grey-out and black-out. For classes such as sunglasses and obelisk in which the shape carries a lot of information, the accuracy drop for blackout or greyout is much smaller than layer masking. This issue is exacerbated by the fact that the baseline grey or black color is relatively close to the true color of the objects for many of these classes, for e.g. obelisks are generally grey, sunglasses are black, etc. In some classes like T-shirt, the accuracy difference can even be negative! This could lead us to \textit{overestimate}  the model's reliance on unmasked features.

Surprisingly, there are a few classes in which the accuracy drop for greyout and blackout is much higher than for layer masking. It turns these classes have other closely related classes which are also associated with grey or black. For example, the priarie grouse (a bird) gets frequently misclassified as the black grouse under black out masking. Similarly, the bee eater (a bird) gets misclassified as a kite or vulture. This could lead us to \textit{underestimate} the model's reliance on unmasked features. Layer masking removes such strong misleading signals, thus the accuracy is higher.

\subsection{Effect of data augmentation on missingness bias}
\label{sec:data_aug_disc}

\begin{figure}[ht]
    \centering
    \includegraphics[width=0.5\textwidth]{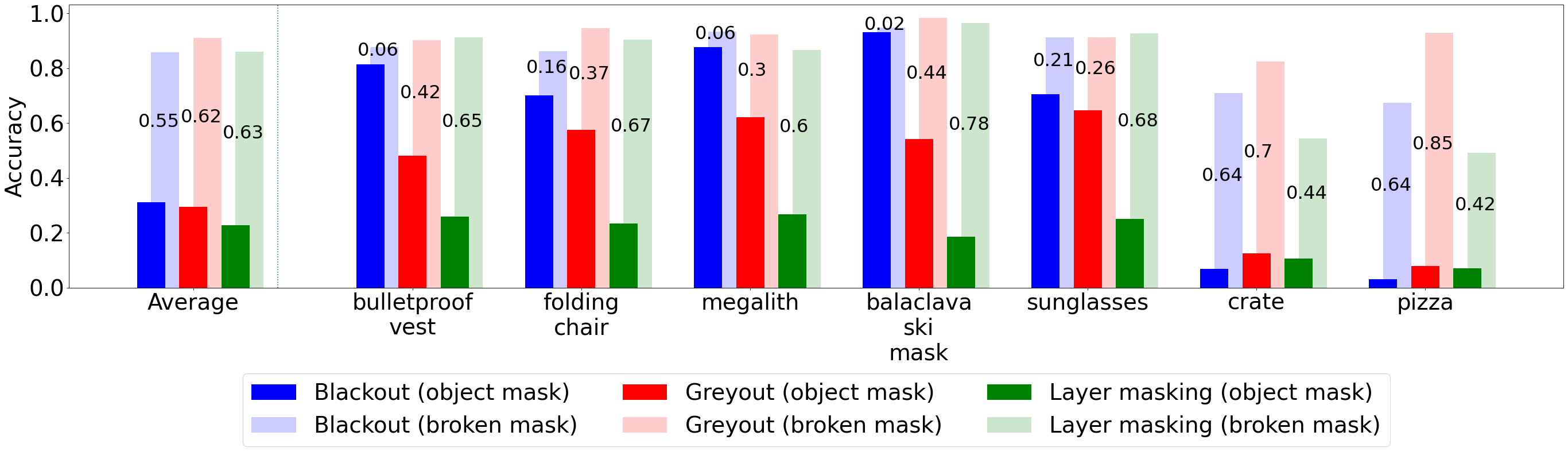}

    \caption{Same as \cref{fig:shape_bias_barplot} but for ResNet-50 pretrained with data augmentations with grey missingness approximations}
    \label{fig:shape_bias_barplot_2}
\end{figure}

Data augmentation strategies like AutoAugment\cite{47890}, RandAugment\cite{cubuk2019randaugment} and RandomErasing \cite{randerase} are sometimes used in the pretraining process for improved performance and generalization. They typically use a grey baseline color as a missingness approximation. Thus, models using these augmentations are very robust to greying out large parts of the input image (see \cref{fig:all_metrics}). However, they may still rely the shape of the mask for their predictions. This issue is not as easily fixable by more training. We evaluate the shape sensitivity of a ResNet-50 trained with data augmentations\cite{wightman2021resnet} as in \cref{sec:shape_bias} and show results in \cref{fig:shape_bias_barplot_2}. We observe that although the average accuracy of greyout/blackout baselines is now comparable to or higher than layer masking, the accuracy drop for layer masking is still a bit higher than that of blackout or greyout. Looking at accuracy over selected classes as before, we find that for some classes like sunglasses and megalith, the accuracy drop is still lower for blackout/greyout which implies that the model is relying on the mask shape to make its predictions. 
However, for classes like crate or pizza, the object mask covers most of the image and its shape does not reveal a lot of information about the class (\cref{fig:example_shape_mask}). 
For these classes, all masking techniques have low accuracies when the object is masked. But when the mask is broken and parts of the object are visible, greyout works best as the model is exposed to grey color frequently during pre-training and is robust to such occlusions.  Thus, the masking technique should be chosen with care depending on the specific image and use case.

\subsection{Impact of masking techniques on LIME }

\begin{table*}[ht!]
    \centering
    \begin{tabular}{@p{24mm}|^p{13mm}^p{10mm}^p{10mm}|^p{13mm}^p{10mm}^p{10mm}|^p{13mm}^p{10mm}^p{10mm}}
     \hline
                  & \multicolumn{3}{c|}{Top-20 ablation accuracy ($\downarrow$)} &  \multicolumn{3}{c|}{Alignment score  ($\uparrow$)} &  \multicolumn{3}{c}{Top-20 Jaccard similarity  ($\uparrow$)}  \\
                  &  Quickshift & $16 \times 16$ & SLIC & Quickshift & $16 \times 16$ & SLIC &  Quickshift & $16 \times 16$ & SLIC\\
                  \hline
        Blackout &  0.570 &  0.701 &  0.740 &  0.138 &  0.020 &  0.092 &  0.186 &  0.087 &  0.131 \\
        Greyout &  0.348 &  0.609 &  0.574 &  0.231 &  0.078 &  0.171 &  0.231 &  0.113 &  0.172 \\
        \rowstyle{\bfseries}  Layer masking &  0.229 &  0.334 &  0.406 &  0.324 &  0.250 &  0.280 &  0.273 &  0.186 &  0.216 \\
         \hline

    \end{tabular}
    \caption{Top-20 ablation accuracy, alignment score, and top-20 Jaccard similarity of LIME scores  over 512 random images }
    \label{tab:lime_metrics}
\end{table*}
We investigate the effect of masking methods on interpretability methods in this section using the example of LIME. Local Interpretable Model-Agnostic Explanations \cite{lime} or LIME is an interpretability method used to explain the predictions of black-box machine learning models by providing locally faithful and human-interpretable explanations. It works by approximating the decision boundary of a model in the vicinity of a particular instance or prediction using a local, interpretable model. The local model is trained on images where the image features are randomly masked out. The weights of the local model represents the importance of each feature in the prediction.

\subsubsection{Visual inspection}

\begin{figure}[ht]

    ~
        \centering
        \fbox{\includegraphics[ width=0.17\textwidth,  height=0.15\textwidth]{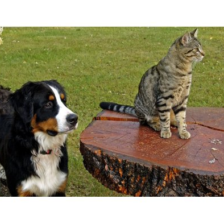}}
        \includegraphics[ width=0.49\textwidth, height=0.3\textwidth]{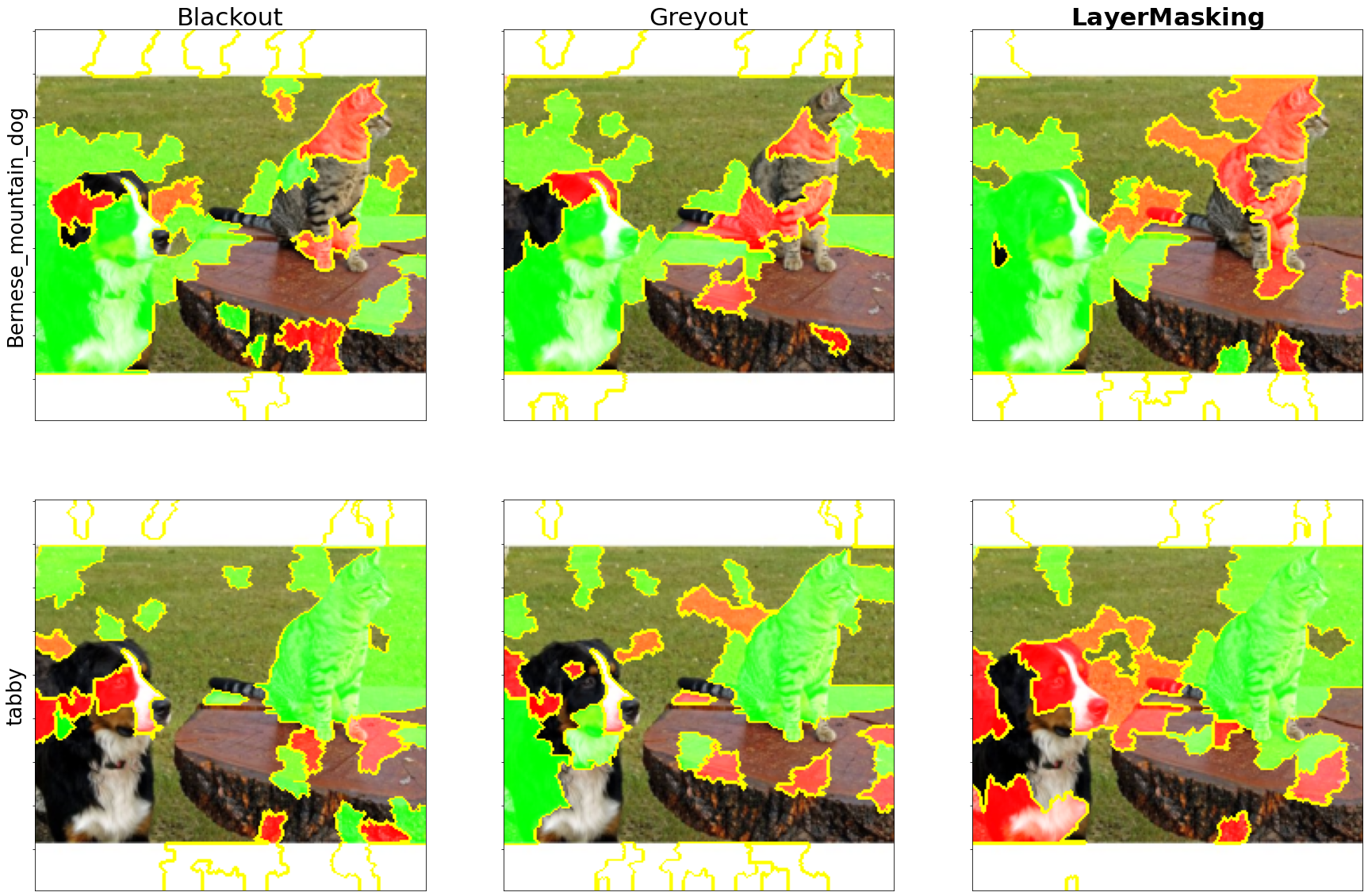}
    \caption{Visualization of LIME scores for the top two predictions of ResNet-50 on a sample image. Columns correspond to the masking techniques (blacking out, greying out, and layer masking), rows are the top 2 predictions. The top two predictions are Bernese mountain dog and tabby cat. Green regions contribute to the prediction, red regions detract from the prediction.}
    \label{fig:lime_examples}
\end{figure}

We first visually illustrate the impact of masking technique on LIME using an example of an image of a cat and a dog, (\cref{fig:lime_examples}). For the LIME explanations, the top 10 segments with the highest magnitude LIME score are highlighted. Red segments detract from the prediction, and a green segments contribute to the prediction. 
 LIME with greyout or blackout masking assigns parts of the cat high positive scores (in green) and parts of the dog negative scores (in red) in the explanation of the prediction of Bernese mountain dog, and vice versa for the explanation of the tabby prediction. The third column containing the LIME explanations obtained using layer masking is much more aligned with human intuition. Visually, LIME with layer masking seems to produce better explanations. We present more examples in the supplementary.

\subsubsection{Quantitative evaluation}

We use these  metrics for evaluating LIME explanations:

(a) \textbf{Top-$k$ ablation test} \cite{7552539, sturmfels2020visualizing}: Choose the $k$ most important segments according to the explanation, remove them by substituting with a missingness approximation (we use grey), and compute the accuracy on the masked images. The more the accuracy drops, the better the explanations. 
 
 (b) \textbf{Alignment score}: Cosine similarity between importance scores returned by LIME and mean normalized segmentation mask. Formally, assuming we have access to segmentation mask $\vm \in [0, 1]^{d \times d}$ for an image of dimension $d$, we compute  $\vg_i = \sum_{(u, v) \in \text{segment } i} \left(\vm[u, v] -\bar{\vm} \right)$ for each segment $i$ where $\bar{\vm}$ is the mean of the segmentation mask. If the importance scores are $\vs$, then the alignment score is the cosine similarity between $\vg$ and $\vs$

(c) \textbf{Top-$k$ Jaccard similarity}: Jaccard similarity between the top $k$ features and the segmentation mask $\vm$. 

While (a) does not depend on any ``ground truth" for evaluation purposes, (b) and (c) use the object segmentation mask as a substitute for the ground truth. We use 512 random images and segmentation masks from Pixel ImageNet \cite{zhang2020interactive} for calculating the above metrics. In these images, the correct class is within the top 3 predictions of the model. We find that layer masking is much superior to blacking or greying out the input across different segmentation algorithms and different metrics, which confirms our intuitions from the visual inspection. As before, the improvement is most significant when the segments are $16 \times 16$ sized patches.

\section{Conclusion}
\label{sec:conc}
In this paper, we have presented a new masking technique such that the model output is both (a) perfectly insensitive to the masked out portion of the input and (b) only focused on the unmasked input and not the masking pattern. We find that layer masking can make CNNs like ResNets very robust to removal of large parts of input without retraining, especially when the masking patterns can get confused with output classes like $16 \times 16$ patch occlusions. Layer masking also does not depend on the shape of the mask, which can be important if we need to precisely mask out only a specific object from the image and the shape of the object carries some useful signal.  We further find that LIME scores obtained using layer masking are better compared to blacking or greying out on multiple metrics like top-k ablation test, top-k Jaccard similarity, and alignment score. We show that this technique can be of great use in both manual feature/object removal for model debugging, and for interpretability techniques like LIME which rely on the ability to remove features from the input without any major distribution shift.

\section{Acknowledgements}

This project was supported in part by a grant from an NSF CAREER AWARD 1942230, ONR YIP award N00014-22-1-2271, ARO's Early Career Program Award 310902-00001, Meta grant 23010098, HR00112090132 (DARPA/RED), HR001119S0026 (DARPA/GARD), Army Grant No. W911NF2120076, NIST 60NANB20D134, the NSF award CCF2212458, an Amazon Research Award and an award from Capital One.

{\small
\bibliographystyle{ieee_fullname}
\bibliography{egbib}
}

\newpage

\onecolumn

\setcounter{section}{0}

\begin{center}
      \Large\textbf{Appendix}\\
   \end{center}

\section{Implementation details}

In order to fairly compare the masking techniques, we fix the number of segments that the segmentation algorithm partitions the image into to approximately equal around 200. We use the \texttt{sklearn} implementation for SLIC and quickshift. For SLIC, we fix the approximate number of segments to 196. For quickshift, we set \texttt{kernel\_size=2}, \texttt{max\_dist=200} , \texttt{ratio=0.2}, which produces approximately 200 segments per image. For LIME, we use 500 random samples to train the linear classifier.  

For the token dropping variant of Vision Transformers (ViT and DeiT), we use code from \url{https://github.com/MadryLab/missingness}.

\section{Comparison of layer masking with partial convolution}

\begin{figure}[ht]
    \centering

    \includegraphics[width=0.45\textwidth]{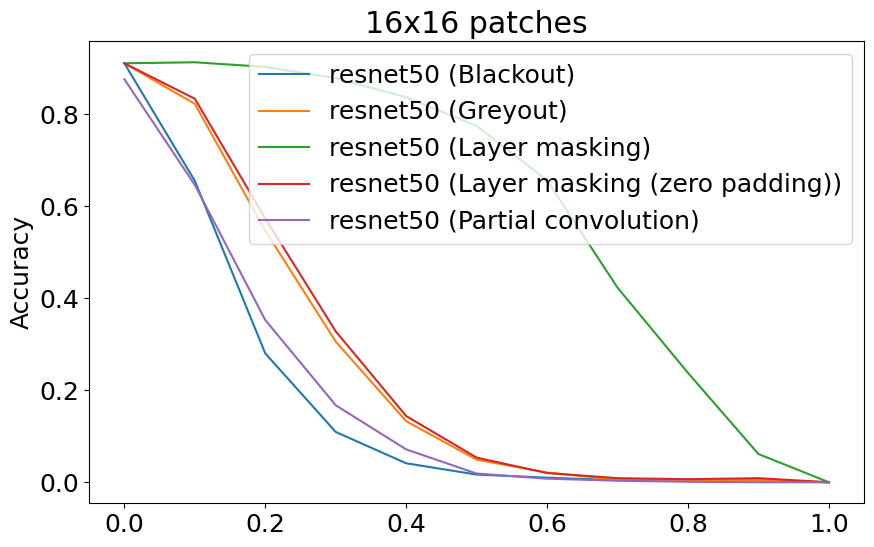}
    \includegraphics[width=0.45\textwidth]{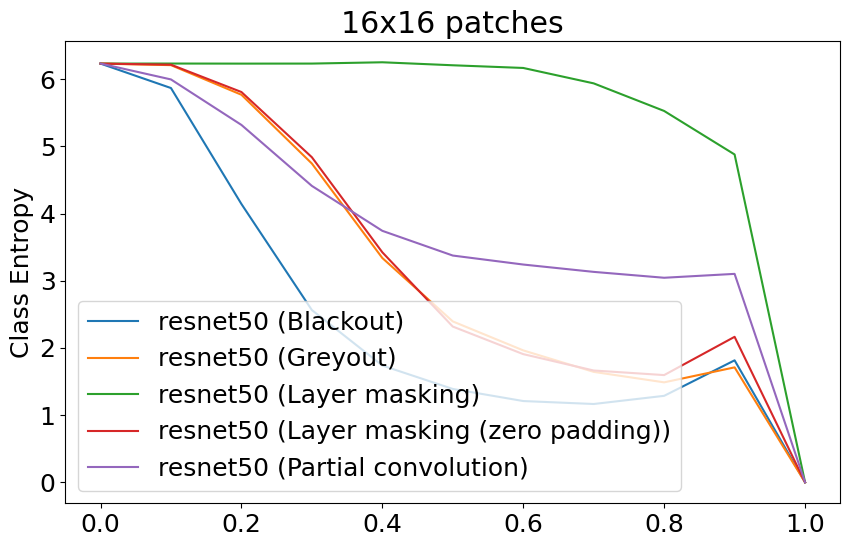}
    \caption{Accuracy and class entropy vs fraction of $16 \times 16$ patches of the image masked out in random order using various masking methods on ResNet-50}
    \label{fig:partialconv}
\end{figure}

Partial convolution is a method for image inpainting introduced by Liu et al, 2018. Partial convolution handles convolution over images with irregular holes by using a method similar to layer masking. However, instead of doing neighbor padding as in layer masking, the convolutions over the edge is \textbf{scaled up} by a factor of $\frac{k^2}{\vm \odot \1_{k\times k}}$ (where $\vm$ is the binary mask corresponding to the field of the convolution and $k$ is the size of the filter). This means that the edge convolutions are given a \textit{higher} weight than normal. While this may be useful for inpainting purposes, where most of the important information is concentrated around the edges and parameters of the neural network can be trained, it is exactly the opposite of what we want, as this worsens the edge artifact problem which we cannot fix by training.  Thus, naively using partial convolution is worse than even zero padding as far as accuracy or unchanged predictions are concerned. We thus find that the AUC for the accuracy (or class entropy) vs fraction of masked image is only \textbf{0.1922} (or \textbf{3.8589})  when we use partial convolution layers, which is much lower than corresponding numbers for layer masking (see \cref{fig:partialconv}).

\clearpage
\section{Ablation study}

To further investigate the effect of layer masking and neighbor padding on model behavior, we construct 3 variants of layer masking: (a) With zero padding instead of neighbor padding  (b)  Masking and padding only the first two residual blocks, (c) Masking and padding only the first convolutional layer, ReLU and BatchNorm layer %

Using a similar setup as in Sec. 4.1, we compute the area under curve (AUC) for each plot of metric vs fraction of segments dropped. The AUC values are averaged over different segmentation algorithms (SLIC, quickshift, etc) and masking orders (random, salient first, etc)(refer \cref{tab:ablation_auc}).

We find that both neighbor padding and masking all layers are important to the masking technique. Layer masking with zero padding is still better than blackout or greyout, but much worse than with neighbor padding. Layer masking only the first two residual blocks is also inferior to masking through all layers, but we find that there are diminishing returns, as we are able to obtain much of the improvement by masking only half of the layers.

\begin{table}[ht]
    \centering
    \resizebox{0.5\textwidth}{!}{%
    \begin{tabular}{p{25mm}p{12mm}p{9mm}p{12mm}p{15mm}}
\toprule
& Accuracy & Class  & Wordnet & Unchanged  \\
&          & Entropy & Similarity & Predictions\\
\midrule
Blackout                                &   0.3881 &        4.6473 &             0.6930 &                0.4094 \\
Greyout                                 &   0.4398 &        4.9408 &             0.7167 &                0.4636 \\
\hline 
Layer masking:&  & & & \\
  \textbf{On all layers }          &   \textbf{0.5604} &        \textbf{5.6021} &             \textbf{0.7881} &                \textbf{0.5907} \\
         
 On 1st and 2nd residual blocks  &   0.5103 &        5.0962 &             0.7616 &                0.5391 \\
 Zero padding            &   0.4502 &        5.0388 &             0.7262 &                0.4747 \\
\bottomrule

\end{tabular}
}
\caption{Average AUC for different variants of layer masking alongside the black out and grey out baselines (model: ResNet-50). Higher the better
\vspace{-15pt}
}
    \label{tab:ablation_auc}
\end{table}

\newpage
\section{Extended results for segment masking experiments (Section 4.1)}

We also measure the degradation of WordNet similarity and change in predictions as segments are removed for models like ResNet-50, ResNet-50 with augmentations, DenseNet, SqueezeNet, AlexNet, EfficientNet and MobileNet. Note that EfficientNet and MobileNet are also trained with grey missingness data augmentations, thus greyout is disproportionately more robust for these models. Most significant differences are found in random $16 \times 16$ patch removal

\begin{table*}[ht]
    \centering
    \begin{tabular}{@p{24mm}|^p{10mm}^p{13mm}^p{13mm}|^p{10mm}^p{13mm}^p{13mm}|^p{10mm}^p{13mm}^p{13mm}}
    \hline
          \textbf{ResNet-50}          & \multicolumn{3}{c|}{Quickshift segments} &  \multicolumn{3}{c|}{$16 \times 16$ patches} &  \multicolumn{3}{c}{SLIC superpixels}  \\
                  &  Random & Most sal. first & Least sal. first &  Random & Most sal. first & Least sal. first &  Random & Most sal. first & Least sal. first\\
                  \hline
     \hspace{-5pt}Unchanged preds &&&&&&&&&\\
        Blackout & 0.415 &  0.134 &  0.832 &  0.193 &  0.359 &  0.608 &  0.362 &  0.213 &  0.699 \\
Greyout &  0.484 &  0.146 &  0.853 &  0.253 &  0.393 &  0.648 &  0.437 &  0.247 &  0.746 \\
\rowstyle{\bfseries} Layer masking &  0.580 &  0.169 &  0.877 &  0.608 &  0.511 &  0.741 &  0.569 &  0.311 &  0.808 \\
         \hline
        \hspace{-5pt}Wordnet Sim &&&&&&&&&\\
       Blackout &  0.705 &  0.547 &  0.889 &  0.571 &  0.672 &  0.802 &  0.669 &  0.591 &  0.838 \\
Greyout &  0.748 &  0.517 &  0.892 &  0.604 &  0.696 &  0.821 &  0.718 &  0.606 &  0.857 \\
\rowstyle{\bfseries} Layer masking  &  0.785 &  0.549 &  0.904 &  0.800 &  0.757 &  0.865 &  0.779 &  0.642 &  0.884 \\
         
                  \hline
     \hspace{-5pt}Accuracy &&&&&&&&&\\
        Blackout &  0.395 &  0.124 &  0.767 &  0.181 &  0.340 &  0.582 &  0.347 &  0.200 &  0.657 \\
        Greyout &  0.463 &  0.137 &  0.787 &  0.240 &  0.374 &  0.621 &  0.418 &  0.234 &  0.702 \\
        \rowstyle{\bfseries} Layer masking &  0.551 &  0.159 &  0.806 &  0.577 &  0.484 &  0.703 &  0.542 &  0.294 &  0.756 \\
         \hline
        \hspace{-5pt}Class entropy &&&&&&&&&\\
       Blackout &  4.988 &  2.224 &  5.824 &  2.574 &  4.570 &  5.406 &  4.815 &  3.976 &  5.661 \\
Greyout &  5.327 &  2.362 &  5.875 &  3.289 &  4.807 &  5.642 &  5.022 &  4.229 &  5.808 \\
 \rowstyle{\bfseries} Layer masking &  5.698 &  2.572 &  5.892 &  5.651 &  5.782 &  5.879 &  5.607 &  4.763 &  5.876 \\
         \hline
    \end{tabular}
    \caption{ AUC of \textbf{(From top)} Fraction of unchanged predictions, Wordnet similarity of the predictions to true label, the accuracy of predictions, and class entropy of the predictions vs fraction of segments masked out for plain ResNet-50 }
    \label{tab:seg_ablate_auc_9}
\end{table*}

\begin{table*}[ht]
    \centering
    \begin{tabular}{@p{24mm}|^p{10mm}^p{13mm}^p{13mm}|^p{10mm}^p{13mm}^p{13mm}|^p{10mm}^p{13mm}^p{13mm}}
    \hline
          \textbf{ResNet-50 (augmented)}          & \multicolumn{3}{c|}{Quickshift segments} &  \multicolumn{3}{c|}{$16 \times 16$ patches} &  \multicolumn{3}{c}{SLIC superpixels}  \\
                  &  Random & Most sal. first & Least sal. first &  Random & Most sal. first & Least sal. first &  Random & Most sal. first & Least sal. first\\
                  \hline
     \hspace{-5pt}Unchanged preds &&&&&&&&&\\
       Blackout      &  0.552 &  0.169 &  0.872 &  0.671 &  0.632 &  0.831 &  0.513 &  0.297 &  0.791 \\
Greyout       &  0.734 &  0.232 &  0.894 &  0.746 &  0.644 &  0.838 &  0.708 &  0.417 &  0.854 \\
Layer masking &  0.621 &  0.186 &  0.884 &  0.622 &  0.557 &  0.775 &  0.603 &  0.348 &  0.828 \\

         \hline
        \hspace{-5pt}Wordnet Sim &&&&&&&&&\\
       Blackout      &  0.787 &  0.577 &  0.912 &  0.838 &  0.823 &  0.903 &  0.770 &  0.654 &  0.886 \\
Greyout       &  0.866 &  0.610 &  0.919 &  0.872 &  0.830 &  0.906 &  0.858 &  0.714 &  0.910 \\
Layer masking &  0.789 &  0.493 &  0.904 &  0.795 &  0.761 &  0.862 &  0.782 &  0.624 &  0.887 \\

                  \hline
     \hspace{-5pt}Accuracy &&&&&&&&&\\
        Blackout      &  0.534 &  0.162 &  0.831 &  0.651 &  0.612 &  0.799 &  0.498 &  0.289 &  0.760 \\
Greyout       &  0.708 &  0.225 &  0.851 &  0.723 &  0.624 &  0.807 &  0.687 &  0.405 &  0.821 \\
Layer masking &  0.603 &  0.180 &  0.843 &  0.605 &  0.540 &  0.749 &  0.585 &  0.338 &  0.798 \\

         \hline
        \hspace{-5pt}Class entropy &&&&&&&&&\\
       Blackout      &  5.520 &  2.458 &  5.878 &  5.742 &  5.807 &  5.892 &  5.438 &  4.511 &  5.834 \\
Greyout       &  5.875 &  2.628 &  5.895 &  5.860 &  5.865 &  5.898 &  5.862 &  4.849 &  5.896 \\
Layer masking &  5.603 &  2.258 &  5.893 &  5.554 &  5.630 &  5.789 &  5.554 &  4.305 &  5.887 \\

         \hline
    \end{tabular}
    \caption{ AUC of \textbf{(From top)} Fraction of unchanged predictions, Wordnet similarity of the predictions to true label, the accuracy of predictions, and class entropy of the predictions vs fraction of segments masked out for ResNet-50 trained with data augmentations}
    \label{tab:seg_ablate_auc_3}
\end{table*}

\begin{table*}[ht]
    \centering
    \begin{tabular}{@p{24mm}|^p{10mm}^p{13mm}^p{13mm}|^p{10mm}^p{13mm}^p{13mm}|^p{10mm}^p{13mm}^p{13mm}}
    \hline
          \textbf{WideResNet-50}          & \multicolumn{3}{c|}{Quickshift segments} &  \multicolumn{3}{c|}{$16 \times 16$ patches} &  \multicolumn{3}{c}{SLIC superpixels}  \\
                  &  Random & Most sal. first & Least sal. first &  Random & Most sal. first & Least sal. first &  Random & Most sal. first & Least sal. first\\
                  \hline
     \hspace{-5pt}Unchanged preds &&&&&&&&&\\
Blackout      &  0.445 &  0.129 &  0.862 &  0.226 &  0.401 &  0.666 &  0.389 &  0.227 &  0.731 \\
Greyout       &  0.515 &  0.141 &  0.877 &  0.293 &  0.447 &  0.707 &  0.472 &  0.263 &  0.778 \\
Layer masking &  0.596 &  0.158 &  0.891 &  0.620 &  0.526 &  0.769 &  0.584 &  0.323 &  0.823 \\

         \hline
        \hspace{-5pt}Wordnet Sim &&&&&&&&&\\
Blackout      &  0.718 &  0.544 &  0.906 &  0.602 &  0.695 &  0.826 &  0.693 &  0.603 &  0.855 \\
Greyout       &  0.757 &  0.529 &  0.909 &  0.624 &  0.727 &  0.846 &  0.733 &  0.618 &  0.874 \\
Layer masking &  0.798 &  0.569 &  0.919 &  0.810 &  0.768 &  0.878 &  0.793 &  0.661 &  0.897 \\

                  \hline
     \hspace{-5pt}Accuracy &&&&&&&&&\\
Blackout      &  0.435 &  0.125 &  0.835 &  0.220 &  0.394 &  0.652 &  0.379 &  0.222 &  0.711 \\
Greyout       &  0.504 &  0.137 &  0.851 &  0.288 &  0.440 &  0.694 &  0.461 &  0.258 &  0.758 \\
Layer masking &  0.587 &  0.154 &  0.864 &  0.609 &  0.516 &  0.754 &  0.574 &  0.317 &  0.802 \\

         \hline
        \hspace{-5pt}Class entropy &&&&&&&&&\\
Blackout      &  5.161 &  2.102 &  5.865 &  2.706 &  4.855 &  5.585 &  4.967 &  4.224 &  5.706 \\
Greyout       &  5.372 &  2.177 &  5.884 &  3.299 &  5.204 &  5.718 &  5.165 &  4.408 &  5.815 \\
Layer masking &  5.734 &  2.354 &  5.891 &  5.668 &  5.790 &  5.884 &  5.664 &  4.819 &  5.878 \\

         \hline
    \end{tabular}
    \caption{ AUC of \textbf{(From top)} Fraction of unchanged predictions, Wordnet similarity of the predictions to true label, the accuracy of predictions, and class entropy of the predictions vs fraction of segments masked out for WideResNet-50}
    \label{tab:seg_ablate_auc_4}
\end{table*}

\begin{table*}[ht]
    \centering
    \begin{tabular}{@p{24mm}|^p{10mm}^p{13mm}^p{13mm}|^p{10mm}^p{13mm}^p{13mm}|^p{10mm}^p{13mm}^p{13mm}}
    \hline
          \textbf{AlexNet}          & \multicolumn{3}{c|}{Quickshift segments} &  \multicolumn{3}{c|}{$16 \times 16$ patches} &  \multicolumn{3}{c}{SLIC superpixels}  \\
                  &  Random & Most sal. first & Least sal. first &  Random & Most sal. first & Least sal. first &  Random & Most sal. first & Least sal. first\\
                  \hline
     \hspace{-5pt}Unchanged preds &&&&&&&&&\\
Blackout      &  0.278 &  0.104 &  0.789 &  0.128 &  0.224 &  0.431 &  0.234 &  0.147 &  0.573 \\
Greyout       &  0.364 &  0.111 &  0.830 &  0.197 &  0.268 &  0.501 &  0.326 &  0.183 &  0.663 \\
Layer masking &  0.437 &  0.120 &  0.853 &  0.487 &  0.358 &  0.597 &  0.458 &  0.226 &  0.733 \\

         \hline
        \hspace{-5pt}Wordnet Sim &&&&&&&&&\\
Blackout      &  0.629 &  0.526 &  0.856 &  0.499 &  0.584 &  0.704 &  0.590 &  0.551 &  0.774 \\
Greyout       &  0.681 &  0.523 &  0.874 &  0.536 &  0.615 &  0.745 &  0.652 &  0.577 &  0.815 \\
Layer masking &  0.721 &  0.527 &  0.882 &  0.734 &  0.676 &  0.794 &  0.729 &  0.601 &  0.845 \\

                  \hline
     \hspace{-5pt}Accuracy &&&&&&&&&\\
Blackout      &  0.256 &  0.091 &  0.696 &  0.115 &  0.205 &  0.397 &  0.215 &  0.132 &  0.517 \\
Greyout       &  0.337 &  0.098 &  0.732 &  0.181 &  0.245 &  0.462 &  0.302 &  0.166 &  0.599 \\
Layer masking &  0.405 &  0.107 &  0.753 &  0.449 &  0.334 &  0.547 &  0.424 &  0.208 &  0.662 \\

         \hline
        \hspace{-5pt}Class entropy &&&&&&&&&\\
Blackout      &  4.587 &  1.917 &  5.790 &  2.794 &  4.444 &  5.243 &  4.342 &  3.930 &  5.522 \\
Greyout       &  5.059 &  2.031 &  5.870 &  3.577 &  4.903 &  5.472 &  4.852 &  4.164 &  5.717 \\
Layer masking &  5.406 &  2.230 &  5.889 &  5.196 &  5.439 &  5.690 &  5.360 &  4.432 &  5.825 \\
         \hline
    \end{tabular}
    \caption{ AUC of \textbf{(From top)} Fraction of unchanged predictions, Wordnet similarity of the predictions to true label, the accuracy of predictions, and class entropy of the predictions vs fraction of segments masked out for AlexNet}
    \label{tab:seg_ablate_auc_5}
\end{table*}

\begin{table*}[ht]
    \centering
    \begin{tabular}{@p{24mm}|^p{10mm}^p{13mm}^p{13mm}|^p{10mm}^p{13mm}^p{13mm}|^p{10mm}^p{13mm}^p{13mm}}
    \hline
          \textbf{SqueezeNet}          & \multicolumn{3}{c|}{Quickshift segments} &  \multicolumn{3}{c|}{$16 \times 16$ patches} &  \multicolumn{3}{c}{SLIC superpixels}  \\
                  &  Random & Most sal. first & Least sal. first &  Random & Most sal. first & Least sal. first &  Random & Most sal. first & Least sal. first\\
                  \hline
     \hspace{-5pt}Unchanged preds &&&&&&&&&\\
Blackout      &  0.285 &  0.103 &  0.794 &  0.141 &  0.243 &  0.469 &  0.240 &  0.147 &  0.578 \\
Greyout       &  0.355 &  0.110 &  0.823 &  0.182 &  0.279 &  0.515 &  0.312 &  0.175 &  0.648 \\
Layer masking &  0.408 &  0.113 &  0.844 &  0.544 &  0.362 &  0.592 &  0.405 &  0.203 &  0.694 \\

         \hline
        \hspace{-5pt}Wordnet Sim &&&&&&&&&\\
Blackout      &  0.627 &  0.524 &  0.842 &  0.515 &  0.579 &  0.721 &  0.599 &  0.545 &  0.767 \\
Greyout       &  0.680 &  0.534 &  0.855 &  0.529 &  0.605 &  0.747 &  0.656 &  0.573 &  0.801 \\
Layer masking &  0.694 &  0.483 &  0.853 &  0.750 &  0.666 &  0.775 &  0.689 &  0.568 &  0.810 \\

                  \hline
     \hspace{-5pt}Accuracy &&&&&&&&&\\
Blackout      &  0.251 &  0.083 &  0.645 &  0.117 &  0.207 &  0.416 &  0.208 &  0.124 &  0.488 \\
Greyout       &  0.311 &  0.090 &  0.669 &  0.154 &  0.237 &  0.456 &  0.270 &  0.148 &  0.546 \\
Layer masking &  0.357 &  0.093 &  0.683 &  0.472 &  0.314 &  0.518 &  0.357 &  0.174 &  0.585 \\

         \hline
        \hspace{-5pt}Class entropy &&&&&&&&&\\
Blackout      &  4.593 &  1.832 &  5.801 &  2.360 &  4.379 &  5.183 &  4.410 &  3.667 &  5.491 \\
Greyout       &  5.017 &  1.935 &  5.865 &  2.931 &  4.755 &  5.483 &  4.776 &  3.903 &  5.673 \\
Layer masking &  5.052 &  2.018 &  5.879 &  5.266 &  5.065 &  5.506 &  4.926 &  4.074 &  5.708 \\

         \hline
    \end{tabular}
    \caption{ AUC of \textbf{(From top)} Fraction of unchanged predictions, Wordnet similarity of the predictions to true label, the accuracy of predictions, and class entropy of the predictions vs fraction of segments masked out for SqueezeNet}
    \label{tab:seg_ablate_auc_6}
\end{table*}

\begin{table*}[ht]
    \centering
    \begin{tabular}{@p{24mm}|^p{10mm}^p{13mm}^p{13mm}|^p{10mm}^p{13mm}^p{13mm}|^p{10mm}^p{13mm}^p{13mm}}
    \hline
          \textbf{DenseNet}          & \multicolumn{3}{c|}{Quickshift segments} &  \multicolumn{3}{c|}{$16 \times 16$ patches} &  \multicolumn{3}{c}{SLIC superpixels}  \\
                  &  Random & Most sal. first & Least sal. first &  Random & Most sal. first & Least sal. first &  Random & Most sal. first & Least sal. first\\
                  \hline
     \hspace{-5pt}Unchanged preds &&&&&&&&&\\
Blackout      &  0.423 &  0.124 &  0.852 &  0.184 &  0.387 &  0.629 &  0.373 &  0.217 &  0.709 \\
Greyout       &  0.481 &  0.134 &  0.865 &  0.272 &  0.403 &  0.652 &  0.449 &  0.247 &  0.751 \\
Layer masking &  0.483 &  0.127 &  0.873 &  0.503 &  0.434 &  0.671 &  0.497 &  0.256 &  0.774 \\

         \hline
        \hspace{-5pt}Wordnet Sim &&&&&&&&&\\
Blackout      &  0.712 &  0.543 &  0.888 &  0.555 &  0.686 &  0.805 &  0.686 &  0.600 &  0.838 \\
Greyout       &  0.746 &  0.552 &  0.894 &  0.607 &  0.700 &  0.821 &  0.726 &  0.619 &  0.858 \\
Layer masking &  0.733 &  0.550 &  0.897 &  0.743 &  0.726 &  0.829 &  0.737 &  0.625 &  0.864 \\

                  \hline
     \hspace{-5pt}Accuracy &&&&&&&&&\\
Blackout      &  0.400 &  0.115 &  0.776 &  0.173 &  0.366 &  0.594 &  0.351 &  0.204 &  0.656 \\
Greyout       &  0.456 &  0.124 &  0.790 &  0.256 &  0.381 &  0.618 &  0.423 &  0.232 &  0.696 \\
Layer masking &  0.456 &  0.118 &  0.796 &  0.477 &  0.412 &  0.632 &  0.469 &  0.243 &  0.718 \\

         \hline
        \hspace{-5pt}Class entropy &&&&&&&&&\\
Blackout      &  4.987 &  1.999 &  5.861 &  2.606 &  5.090 &  5.555 &  4.868 &  4.092 &  5.667 \\
Greyout       &  5.321 &  2.160 &  5.880 &  3.549 &  5.176 &  5.660 &  5.234 &  4.365 &  5.798 \\
Layer masking &  5.051 &  2.017 &  5.885 &  5.152 &  5.391 &  5.661 &  5.050 &  4.215 &  5.773 \\

         \hline
    \end{tabular}
    \caption{ AUC of \textbf{(From top)} Fraction of unchanged predictions, Wordnet similarity of the predictions to true label, the accuracy of predictions, and class entropy of the predictions vs fraction of segments masked out for DenseNet}
    \label{tab:seg_ablate_auc_7}
\end{table*}

\begin{table*}[ht]
    \centering
    \begin{tabular}{@p{24mm}|^p{10mm}^p{13mm}^p{13mm}|^p{10mm}^p{13mm}^p{13mm}|^p{10mm}^p{13mm}^p{13mm}}
    \hline
          \textbf{MobileNet-v3}          & \multicolumn{3}{c|}{Quickshift segments} &  \multicolumn{3}{c|}{$16 \times 16$ patches} &  \multicolumn{3}{c}{SLIC superpixels}  \\
                  &  Random & Most sal. first & Least sal. first &  Random & Most sal. first & Least sal. first &  Random & Most sal. first & Least sal. first\\
                  \hline
     \hspace{-5pt}Unchanged preds &&&&&&&&&\\
Blackout      &  0.525 &  0.144 &  0.871 &  0.278 &  0.406 &  0.646 &  0.495 &  0.271 &  0.761 \\
Greyout       &  0.661 &  0.181 &  0.888 &  0.633 &  0.576 &  0.779 &  0.633 &  0.355 &  0.815 \\
Layer masking &  0.516 &  0.135 &  0.873 &  0.619 &  0.441 &  0.644 &  0.531 &  0.271 &  0.767 \\

         \hline
        \hspace{-5pt}Wordnet Sim &&&&&&&&&\\
Blackout      &  0.769 &  0.557 &  0.910 &  0.612 &  0.702 &  0.822 &  0.751 &  0.636 &  0.872 \\
Greyout       &  0.829 &  0.542 &  0.911 &  0.817 &  0.791 &  0.878 &  0.816 &  0.665 &  0.889 \\
Layer masking &  0.753 &  0.531 &  0.907 &  0.804 &  0.715 &  0.813 &  0.764 &  0.623 &  0.870 \\
        
                  \hline
     \hspace{-5pt}Accuracy &&&&&&&&&\\
Blackout      &  0.515 &  0.139 &  0.837 &  0.272 &  0.398 &  0.632 &  0.486 &  0.266 &  0.738 \\
Greyout       &  0.644 &  0.176 &  0.853 &  0.621 &  0.562 &  0.757 &  0.617 &  0.345 &  0.788 \\
Layer masking &  0.504 &  0.131 &  0.838 &  0.605 &  0.430 &  0.627 &  0.520 &  0.265 &  0.744 \\

         \hline
        \hspace{-5pt}Class entropy &&&&&&&&&\\
Blackout      &  5.560 &  2.204 &  5.897 &  4.069 &  5.319 &  5.679 &  5.490 &  4.504 &  5.848 \\
Greyout       &  5.813 &  2.324 &  5.897 &  5.776 &  5.805 &  5.886 &  5.771 &  4.753 &  5.895 \\
Layer masking &  5.336 &  2.152 &  5.887 &  5.532 &  5.309 &  5.540 &  5.341 &  4.406 &  5.801 \\

         \hline
    \end{tabular}
    \caption{ AUC of \textbf{(From top)} Fraction of unchanged predictions, Wordnet similarity of the predictions to true label, the accuracy of predictions, and class entropy of the predictions vs fraction of segments masked out for MobileNet}
    \label{tab:seg_ablate_auc_8}
\end{table*}

\begin{table*}[ht]
    \centering
    \begin{tabular}{@p{24mm}|^p{10mm}^p{13mm}^p{13mm}|^p{10mm}^p{13mm}^p{13mm}|^p{10mm}^p{13mm}^p{13mm}}
    \hline
          \textbf{EfficientNet}          & \multicolumn{3}{c|}{Quickshift segments} &  \multicolumn{3}{c|}{$16 \times 16$ patches} &  \multicolumn{3}{c}{SLIC superpixels}  \\
                  &  Random & Most sal. first & Least sal. first &  Random & Most sal. first & Least sal. first &  Random & Most sal. first & Least sal. first\\
                  \hline
     \hspace{-5pt}Unchanged preds &&&&&&&&&\\
Blackout      &  0.591 &  0.162 &  0.883 &  0.261 &  0.439 &  0.666 &  0.581 &  0.326 &  0.805 \\
Greyout       &  0.729 &  0.204 &  0.896 &  0.686 &  0.605 &  0.804 &  0.723 &  0.411 &  0.849 \\
Layer masking &  0.553 &  0.146 &  0.881 &  0.581 &  0.476 &  0.692 &  0.572 &  0.302 &  0.802 \\

         \hline
        \hspace{-5pt}Wordnet Sim &&&&&&&&&\\
Blackout      &  0.804 &  0.569 &  0.910 &  0.595 &  0.716 &  0.824 &  0.800 &  0.667 &  0.887 \\
Greyout       &  0.860 &  0.594 &  0.915 &  0.842 &  0.807 &  0.891 &  0.859 &  0.709 &  0.903 \\
Layer masking &  0.769 &  0.533 &  0.906 &  0.788 &  0.737 &  0.839 &  0.778 &  0.637 &  0.880 \\

                  \hline
     \hspace{-5pt}Accuracy &&&&&&&&&\\
Blackout      &  0.570 &  0.154 &  0.835 &  0.252 &  0.424 &  0.641 &  0.559 &  0.314 &  0.768 \\
Greyout       &  0.697 &  0.194 &  0.847 &  0.661 &  0.580 &  0.772 &  0.692 &  0.394 &  0.809 \\
Layer masking &  0.532 &  0.139 &  0.834 &  0.561 &  0.459 &  0.666 &  0.549 &  0.290 &  0.766 \\

         \hline
        \hspace{-5pt}Class entropy &&&&&&&&&\\
Blackout      &  5.642 &  2.222 &  5.893 &  3.772 &  5.378 &  5.661 &  5.654 &  4.610 &  5.868 \\
Greyout       &  5.879 &  2.430 &  5.893 &  5.848 &  5.857 &  5.897 &  5.880 &  4.941 &  5.895 \\
Layer masking &  5.356 &  2.099 &  5.891 &  5.410 &  5.464 &  5.682 &  5.367 &  4.390 &  5.844 \\

         \hline
    \end{tabular}
    \caption{ AUC of \textbf{(From top)} Fraction of unchanged predictions, Wordnet similarity of the predictions to true label, the accuracy of predictions, and class entropy of the predictions vs fraction of segments masked out for EfficientNet}
    \label{tab:seg_ablate_auc_2}
\end{table*}

\clearpage
\section{Extended experiments on shape bias (Section 4.2 and Section 4.3)}

\begin{figure}[ht]
    \centering
    \begin{subfigure}{\textwidth}
        \includegraphics[width=\textwidth]{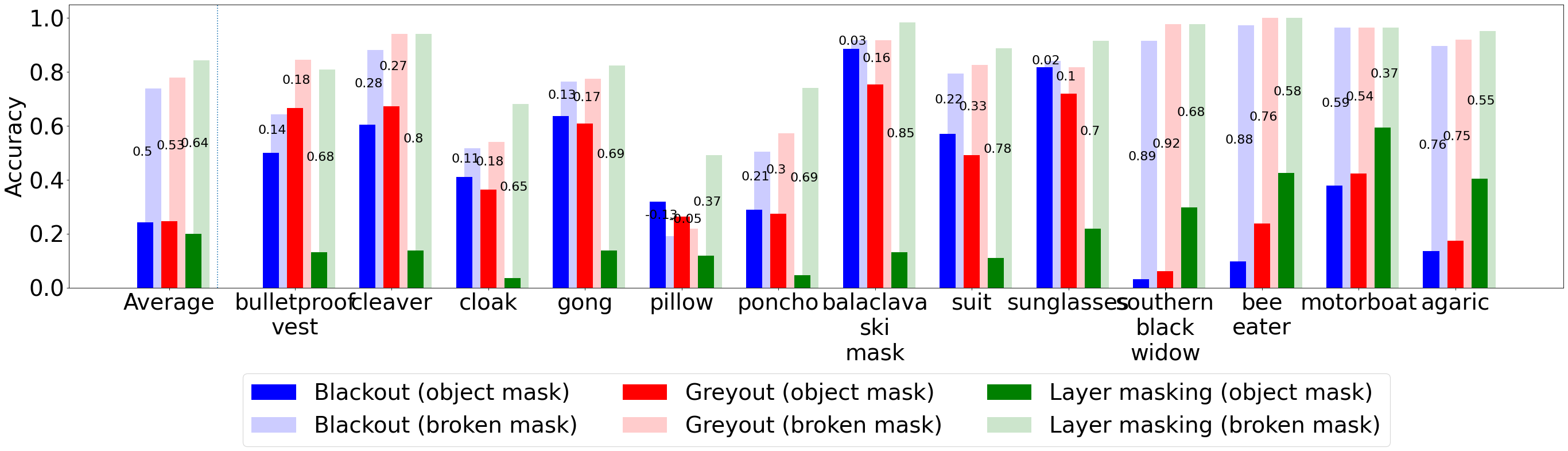}
        \caption{Wide ResNet-50}
    \end{subfigure}
    \begin{subfigure}{\textwidth}
        \includegraphics[width=\textwidth]{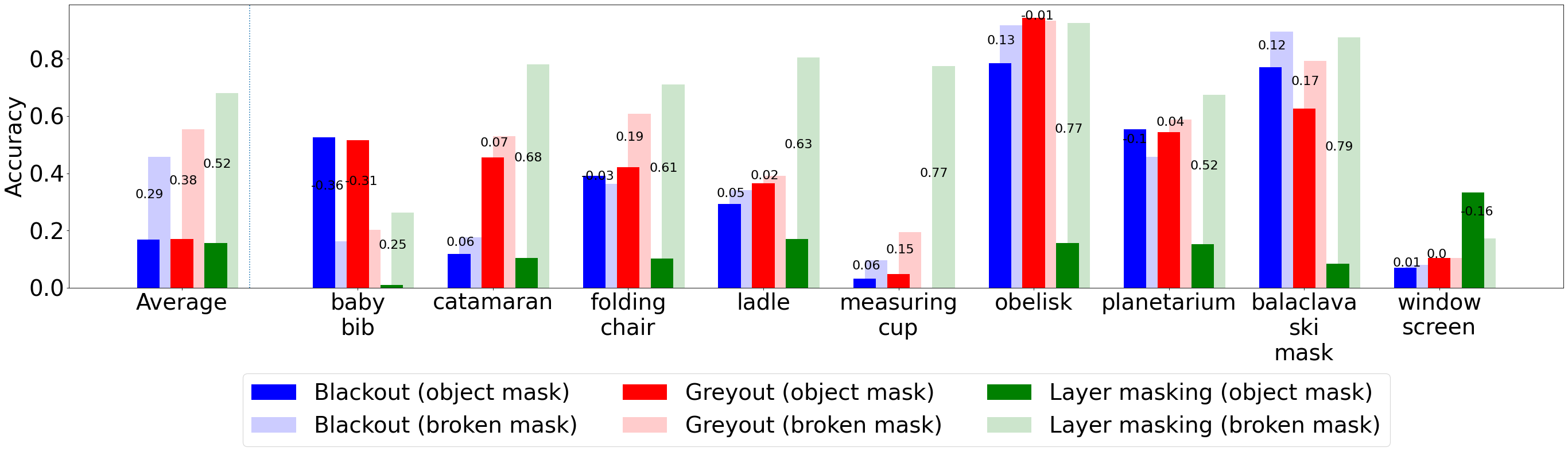}
        \caption{AlexNet}
    \end{subfigure}
    \begin{subfigure}{\textwidth}
        \includegraphics[width=\textwidth]{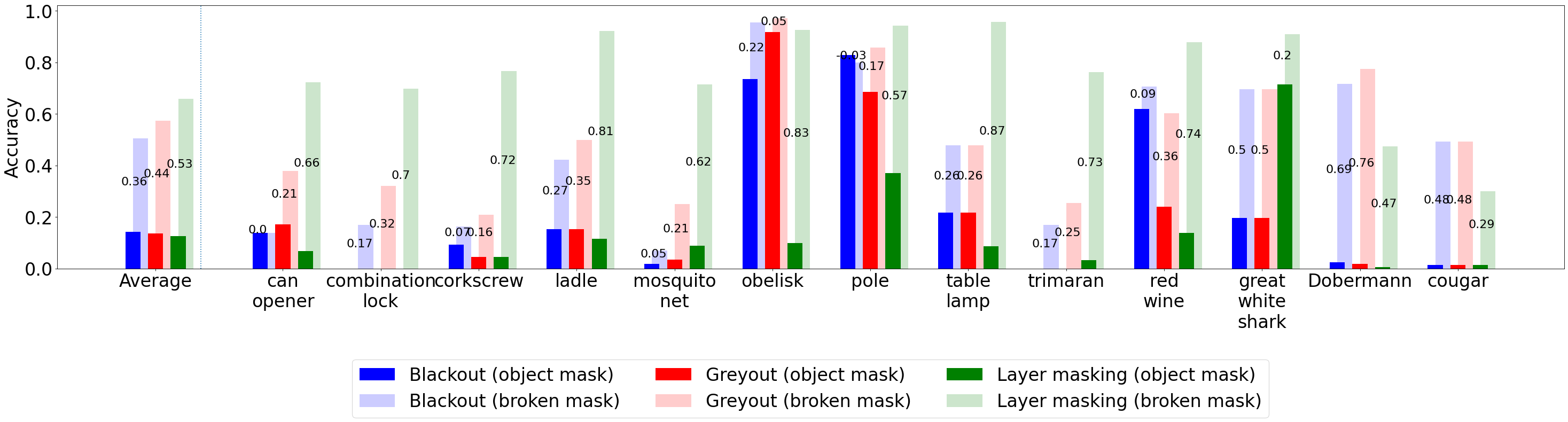}
        \caption{SqueezeNet}
    \end{subfigure}
    \begin{subfigure}{\textwidth}
        \includegraphics[width=\textwidth]{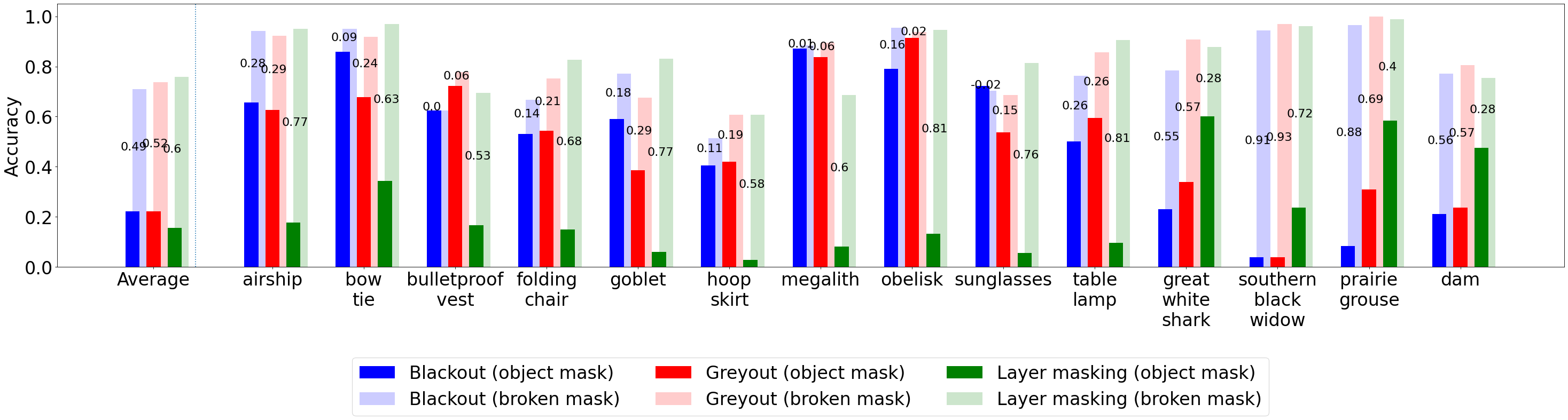}
        \caption{DenseNet}
    \end{subfigure}
    \caption{Effect of shape bias (measured as in Section 4.2) for Wide ResNet-50, AlexNet, SqueezeNet and DenseNet}
    \label{fig:sb_1}
\end{figure}

We now show the bar plots for object masked and broken masked cases for different CNN architectures. Consistent with the previous section, we observe that layer masking is more robust as compared to black-out or grey-out for Wide ResNet-50, AlexNet, SqueezeNet and DenseNet (\cref{fig:sb_1}). Also, the object masked accuracy is typically lower on average. Looking at specific classes, we see similar trends as mentioned in Section 4.2. There are many classes for like megalith, obelisk, sunglasses, etc in which the object's true color is very close to the masking color, and the shape of the object mask conveys a lot of information about the class itself. Conversely, other classes like priarie grouse, bee eater, southern black widow, etc get misclassified as other related classes when masked out using black or grey baseline colors at a higher -than-ideal rate as compared to layer masking.

\begin{figure}[ht]
    \centering
    \begin{subfigure}{\textwidth}
        \includegraphics[width=\textwidth]{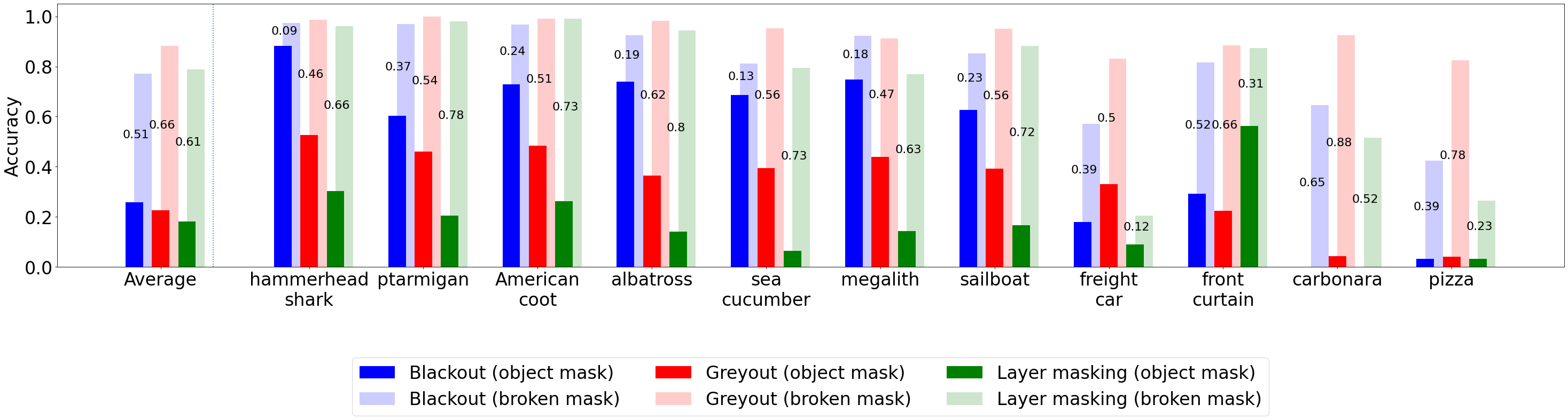}
        \caption{EfficientNet}
    \end{subfigure}
    \begin{subfigure}{\textwidth}
        \includegraphics[width=\textwidth]{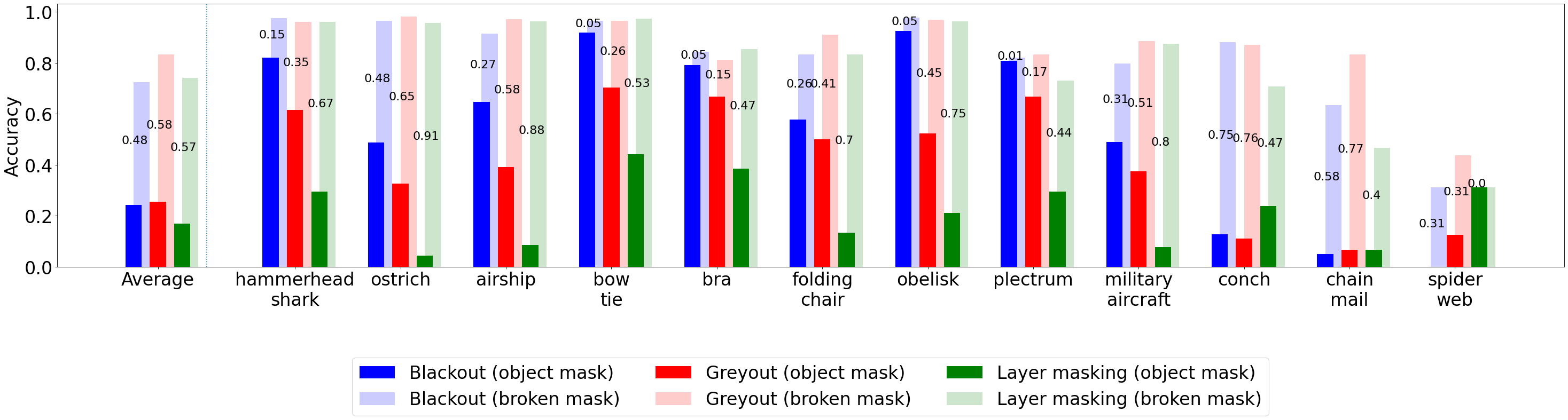}
        \caption{MobileNet}
    \end{subfigure}
    \caption{Effect of shape bias (measured as in Section 4.2) for EfficientNet and MobileNet ()}
    \label{fig:sb_2}
\end{figure}

However, for EfficientNet and MobileNet-v3 (\cref{fig:sb_2}), we find that owing to its pretraining on data augmentations, it is more robust grey-out masking, even compared to layer masking. Still, consistent with Section 4.3, we find classes like megalith and hammerhead shark where layer masking can be more helpful, but also classes like pizza or carbonara where it is not.

In conclusion, we should be cognizant of the missingness biases of a masking method when applied to a model, both shape and color, when evaluating a model's dependence on various image features. Layer masking can be particularly useful in cases where the object to be masked has a distinctive shape with its color also being similar to the baseline color (for e.g: obelisk, megalith, sunglasses). It may also be useful in situations where there exists another class closely related to the true class which has a similar shape but different color which closely resembles the masking color (for e.g: ). It may not be so useful in situations where shape is not very indicative of object and model is already robust to some color replacing masking method like greyout (e.g: pizza, crate, carbonara).

\clearpage
\section{Extended experiments on LIME (Section 4.4)}

\subsection{Qualitative}

\begin{figure}[ht]
    \centering

        \fbox{\includegraphics[ width=0.33\textwidth, height=0.27\textwidth]{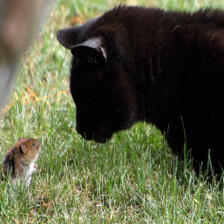}}
        
        \includegraphics[ width=\textwidth, height=0.55\textwidth]{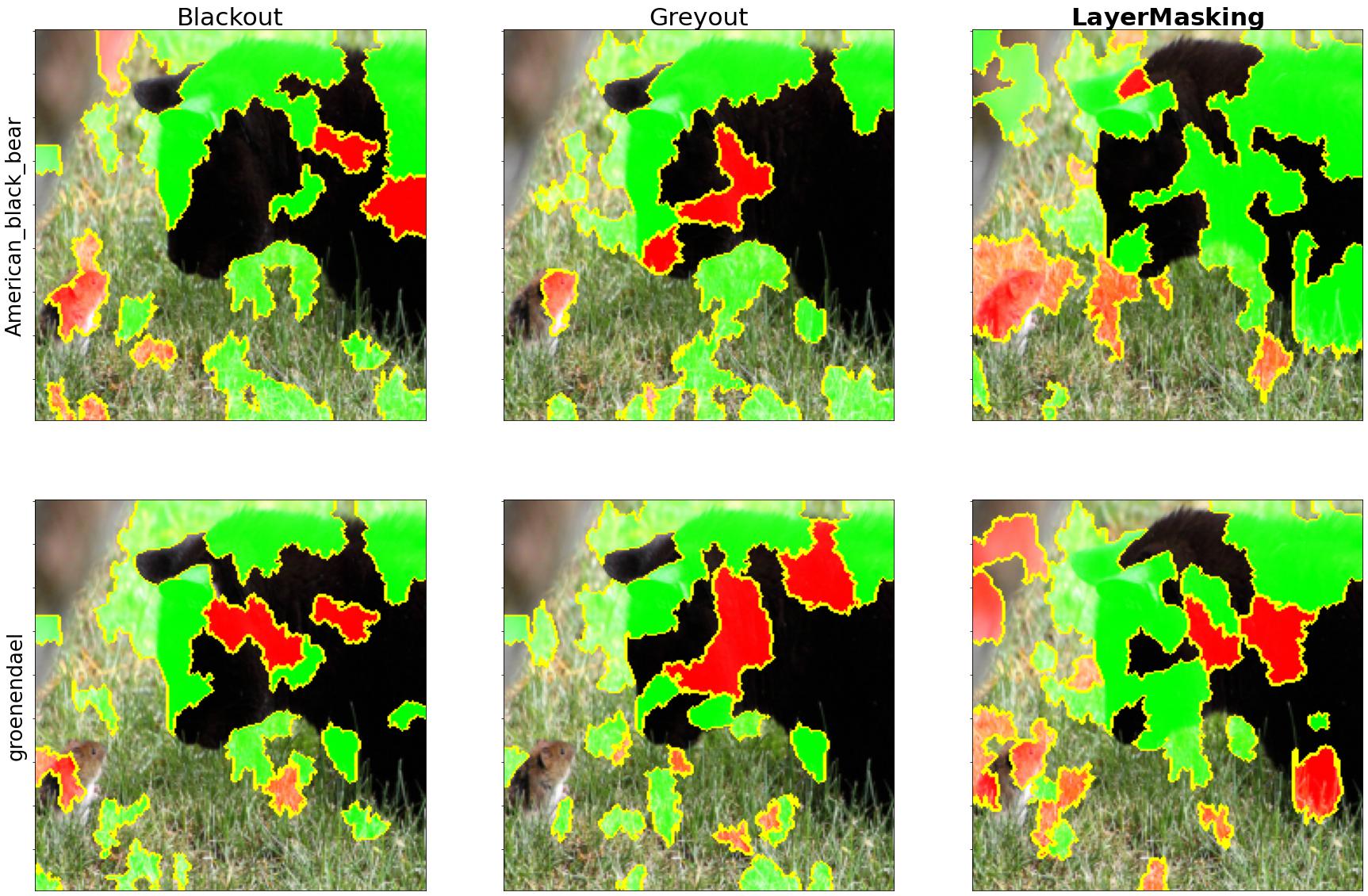}

    \caption{Visualization of LIME scores for the top two predictions of ResNet-50 on a sample image of a cat and a mouse. Columns correspond to the masking techniques (blacking out, greying out, and layer masking), rows are the top 2 predictions. The top two predictions are American black bear and mouse. Green regions contribute to the prediction, red regions detract from the prediction.}
    \label{fig:lime_examples_2}
\end{figure}

We also include some more visualizations of LIME scores on random images from ImageNet, with most important segments highlighted in green (positive score) or red (negative score). These are \textbf{not} cherrypicked.

\begin{figure*}[ht!]
    \centering
    
    \includegraphics[width=0.95\textwidth]{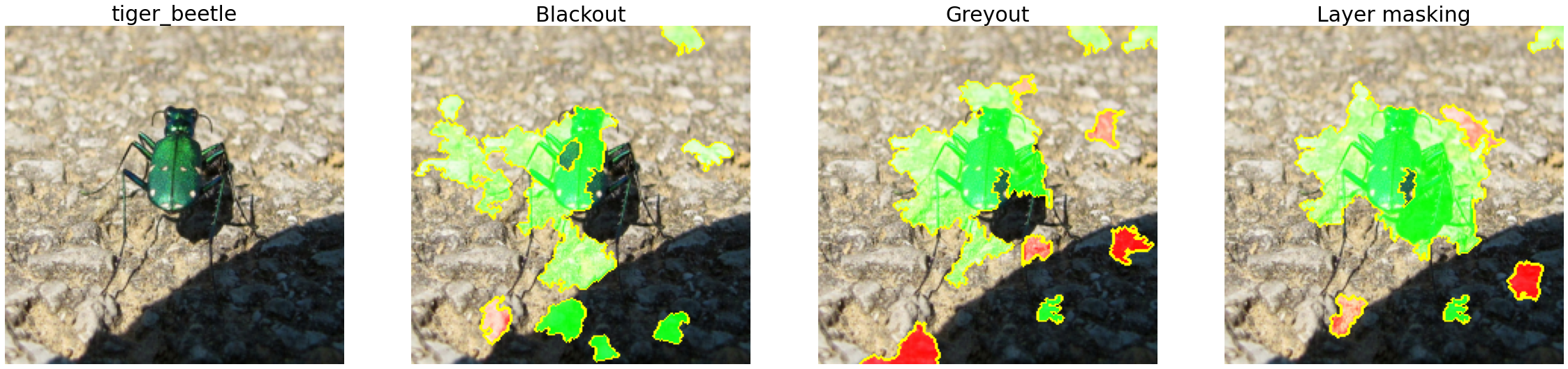}
    \includegraphics[width=0.95\textwidth]{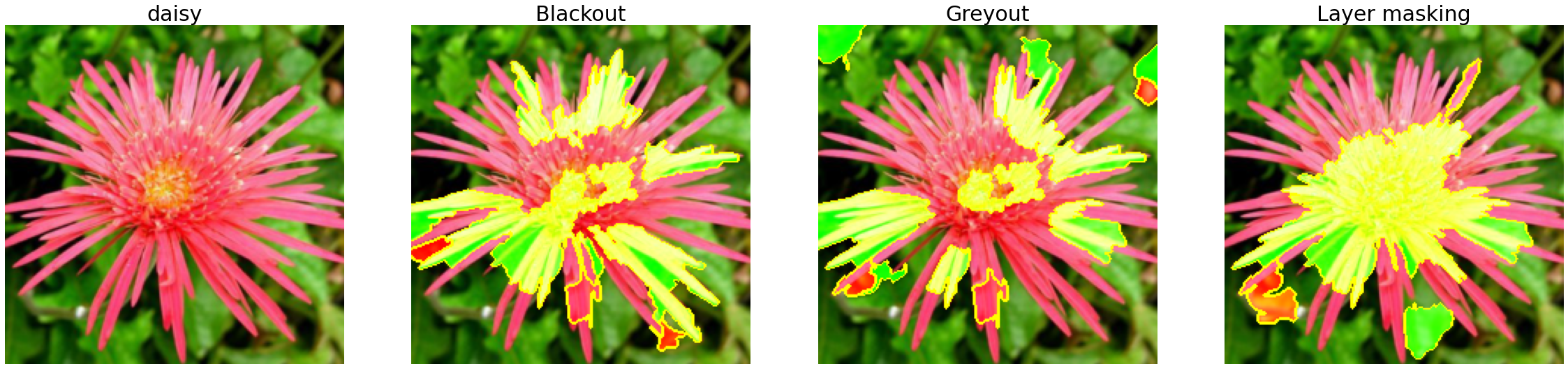}
    \includegraphics[width=0.95\textwidth]{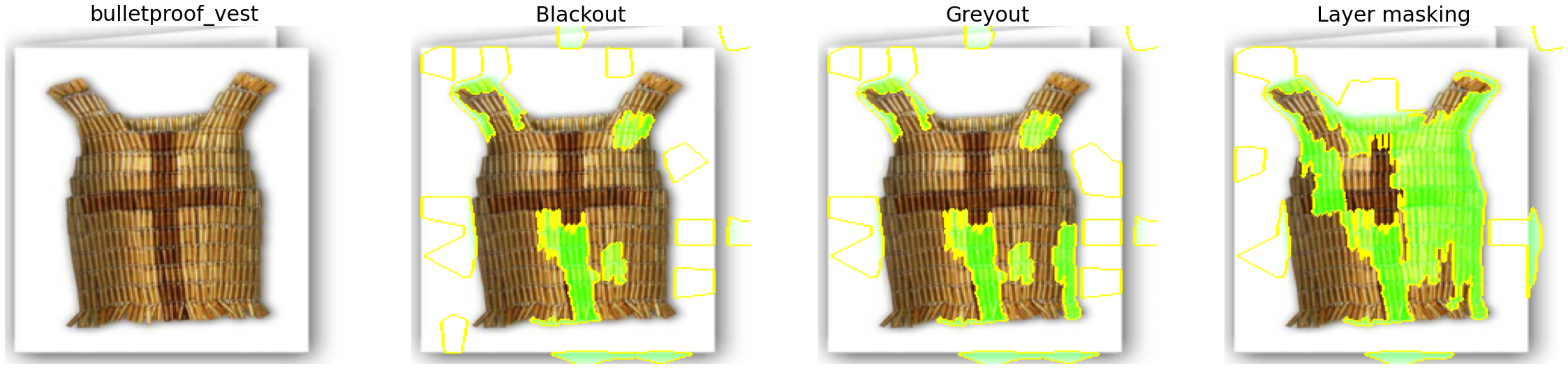}
    \includegraphics[width=0.95\textwidth]{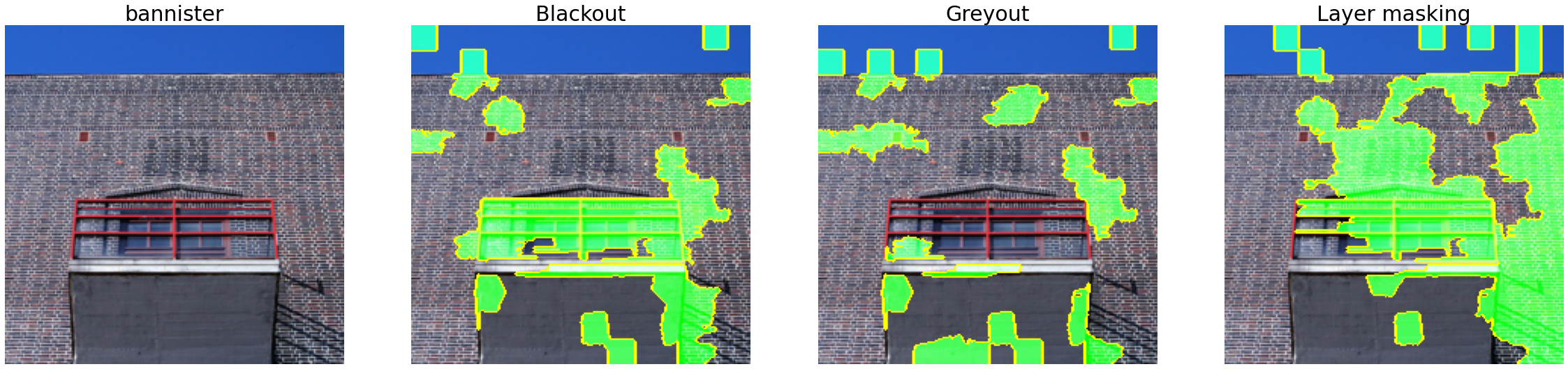}
    \includegraphics[width=0.95\textwidth]{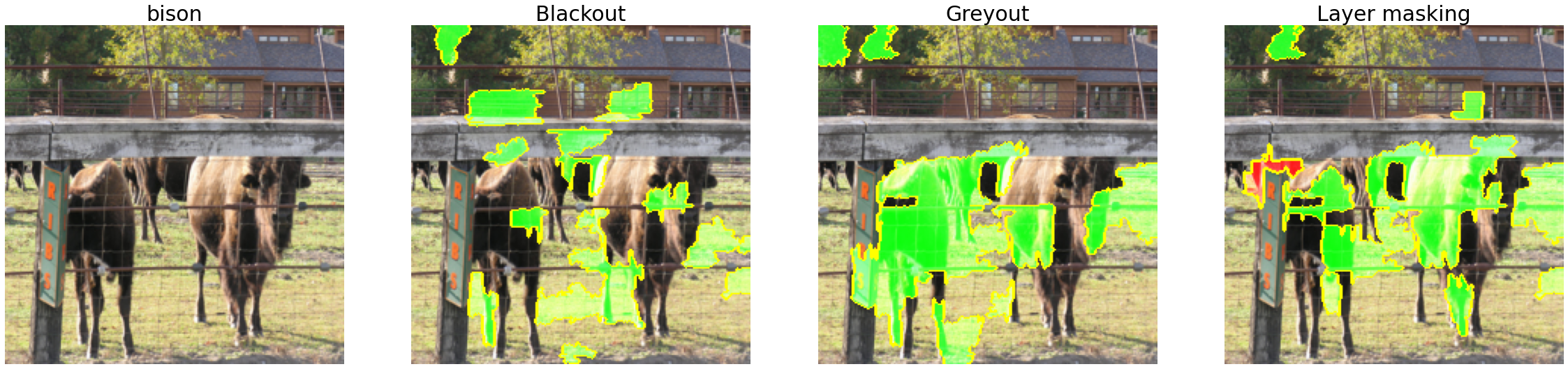}

    \caption{LIME scores using SLIC segmentation (5 samples). Top 15 segments are highlighted}
    \label{fig:lime_slic_ex}
\end{figure*}

\begin{figure*}[ht!]
    \centering

    \includegraphics[width=\textwidth]{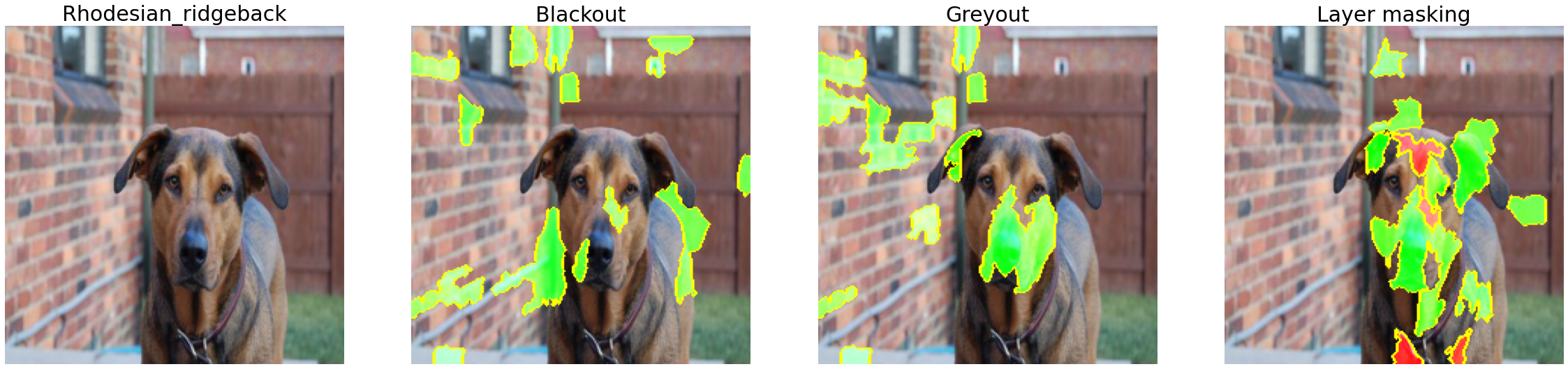}
    \includegraphics[width=\textwidth]{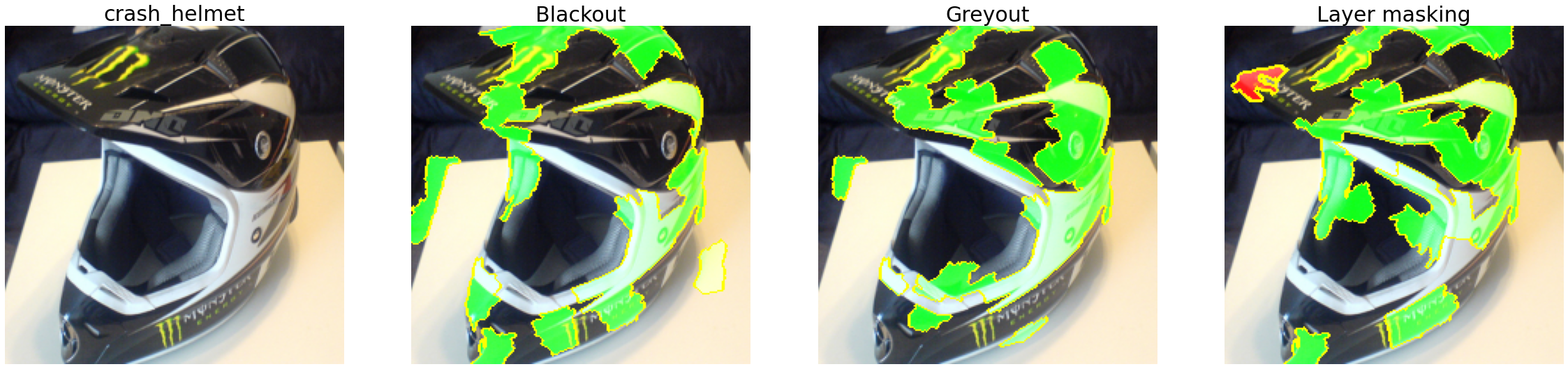}
    \includegraphics[width=\textwidth]{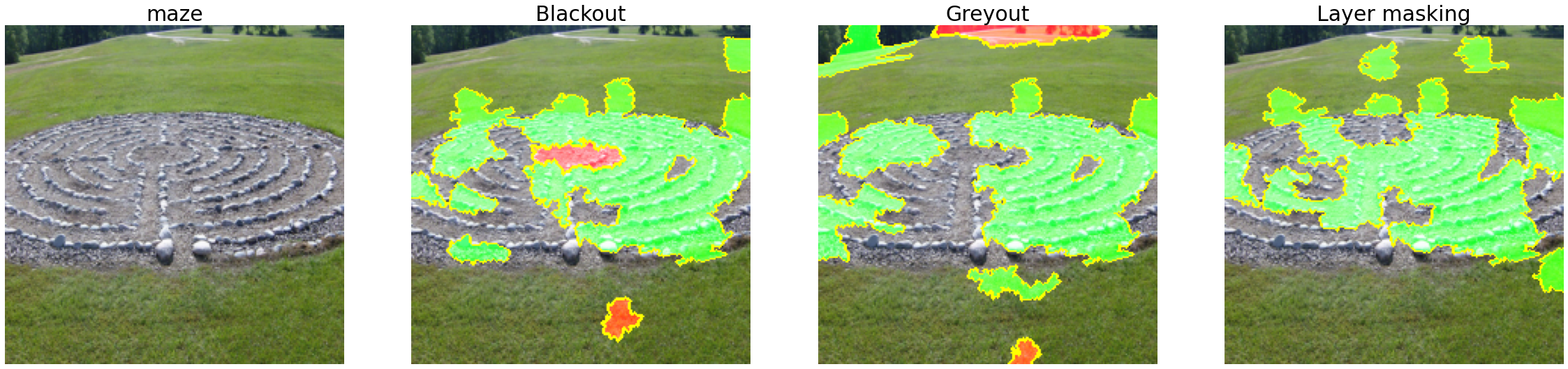}
    \includegraphics[width=\textwidth]{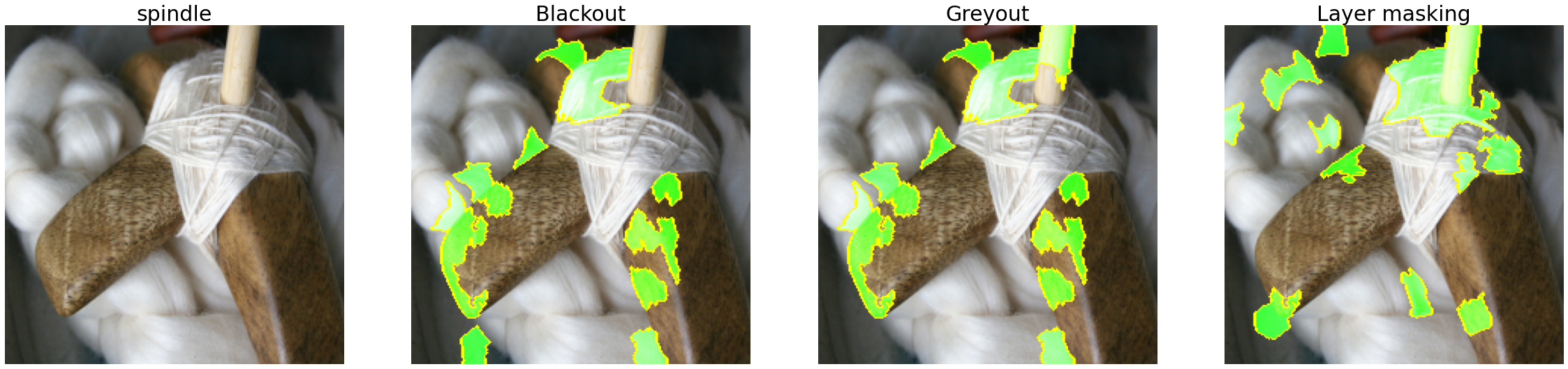}
    \includegraphics[width=\textwidth]{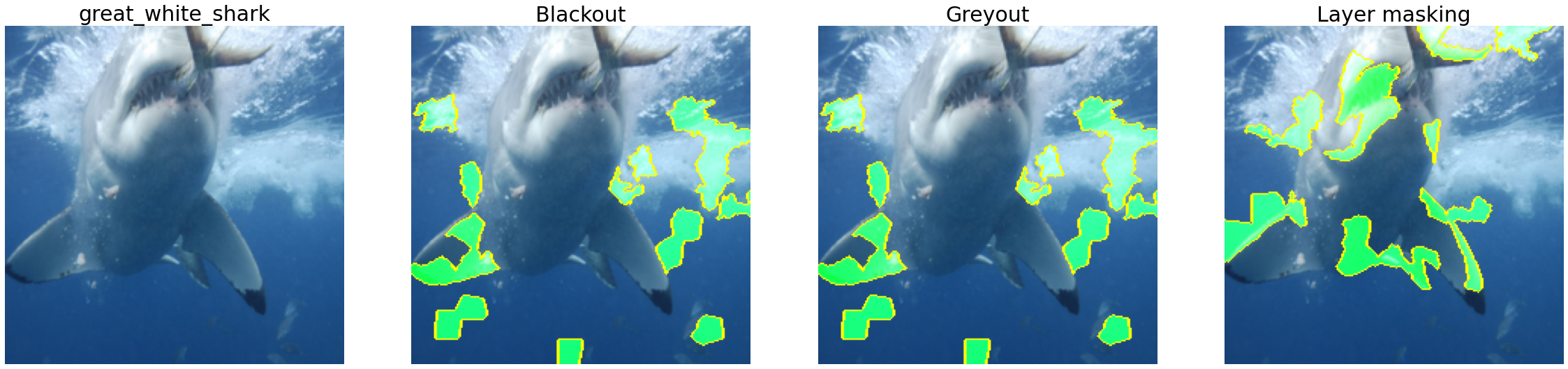}

    \caption{LIME scores using SLIC segmentation (5 samples). Top 15 segments are highlighted}
    \label{fig:lime_slic_ex_2}
\end{figure*}

\begin{figure*}[ht!]
    \centering
    
    \includegraphics[width=\textwidth]{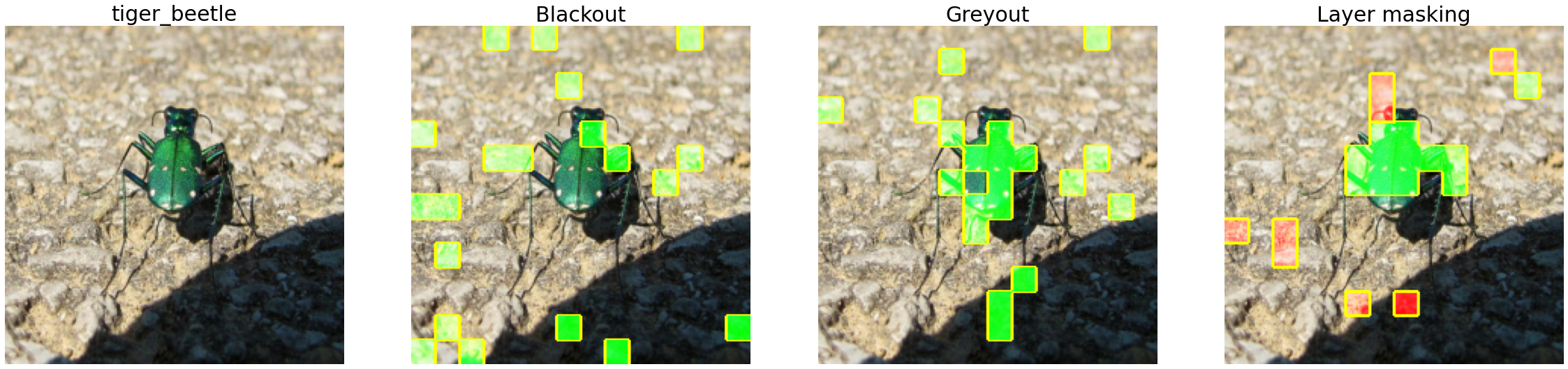}
    \includegraphics[width=\textwidth]{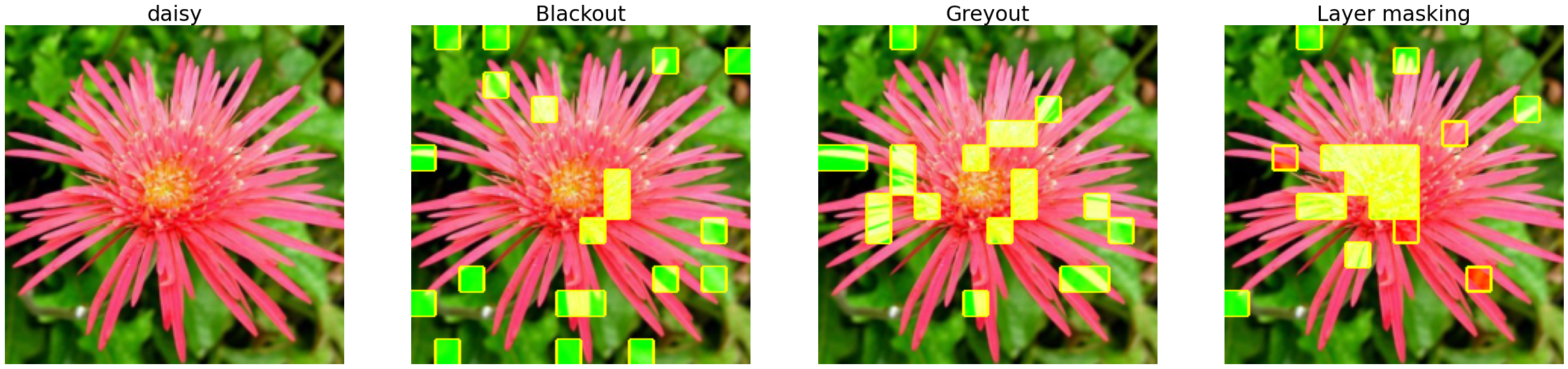}
    \includegraphics[width=\textwidth]{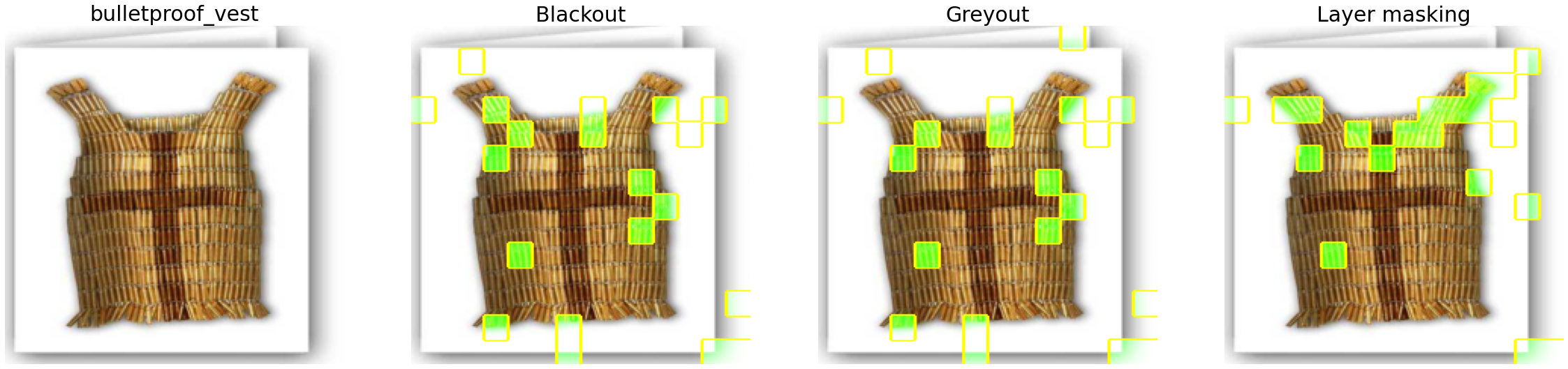}
    \includegraphics[width=\textwidth]{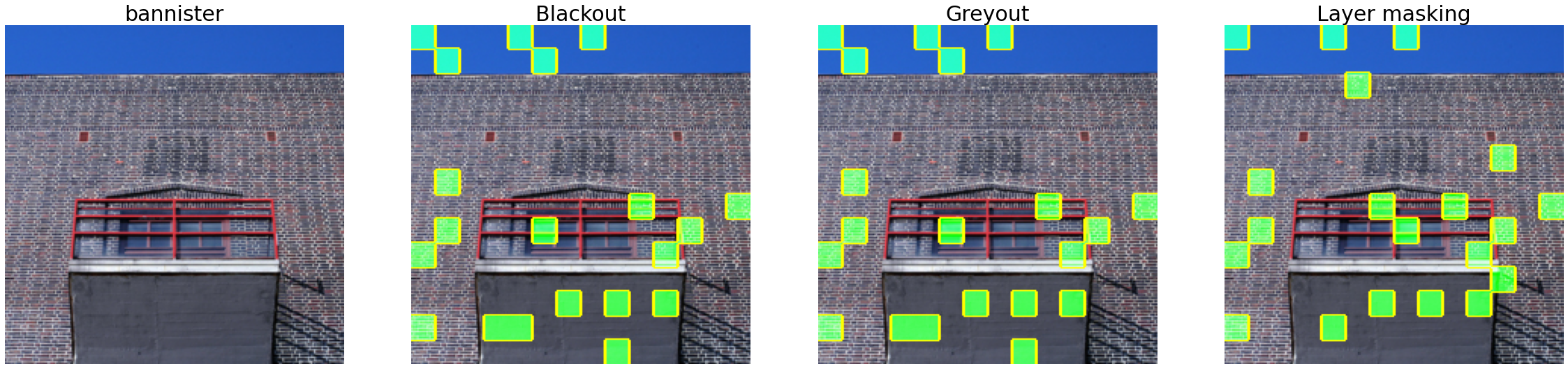}
    \includegraphics[width=\textwidth]{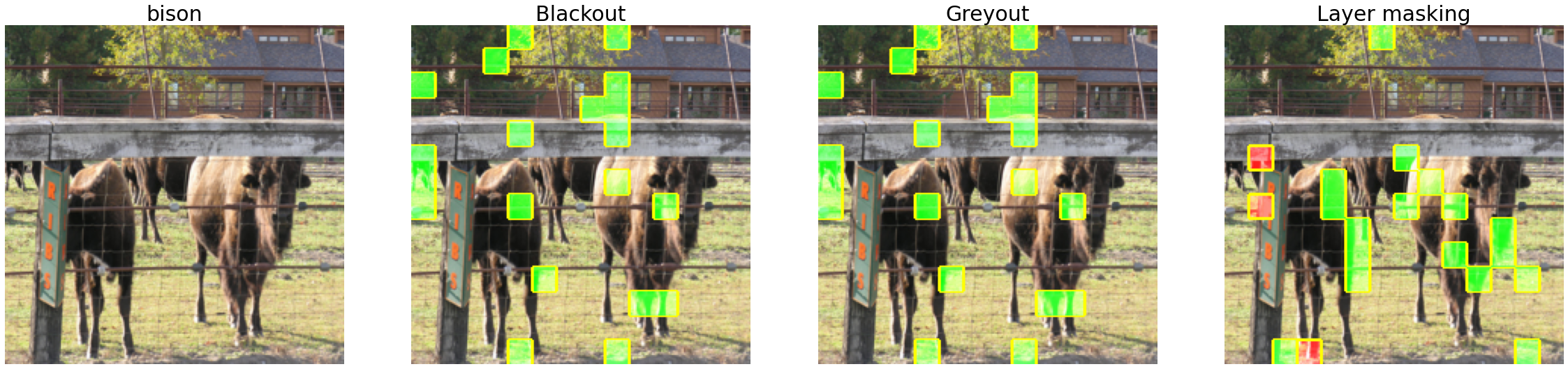}

    \caption{LIME scores using $16 \times 16$ segmentation (5 samples). Top 20 segments are highlighted}
    \label{fig:lime_16_ex}
\end{figure*}

\begin{figure*}[ht!]
    \centering

    \includegraphics[width=\textwidth]{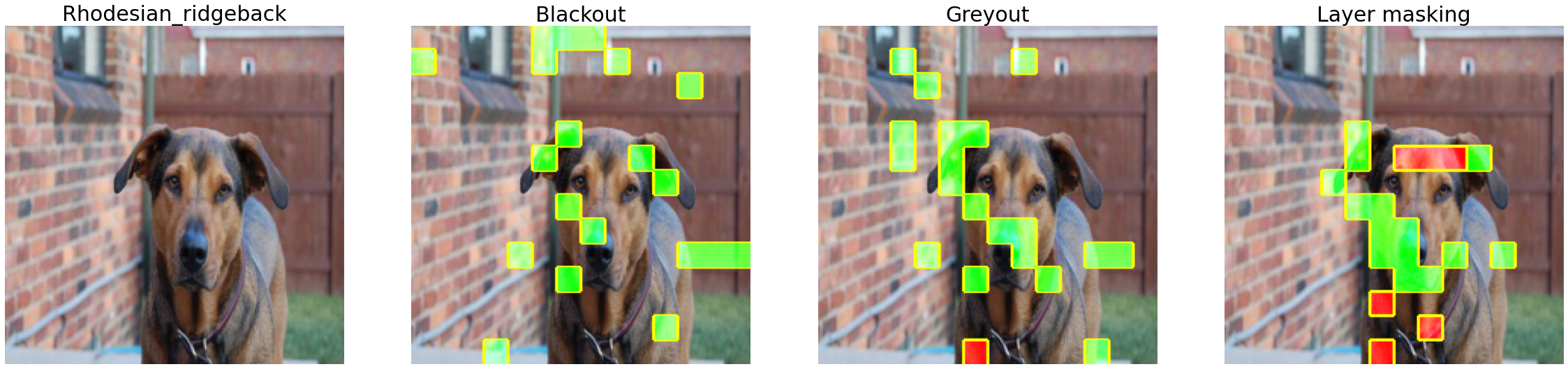}
    \includegraphics[width=\textwidth]{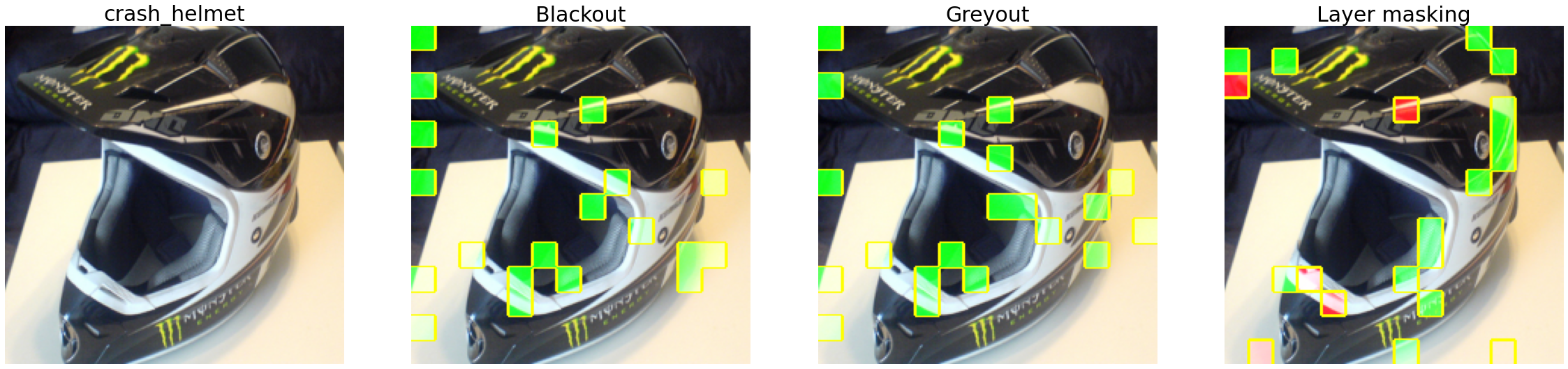}
    \includegraphics[width=\textwidth]{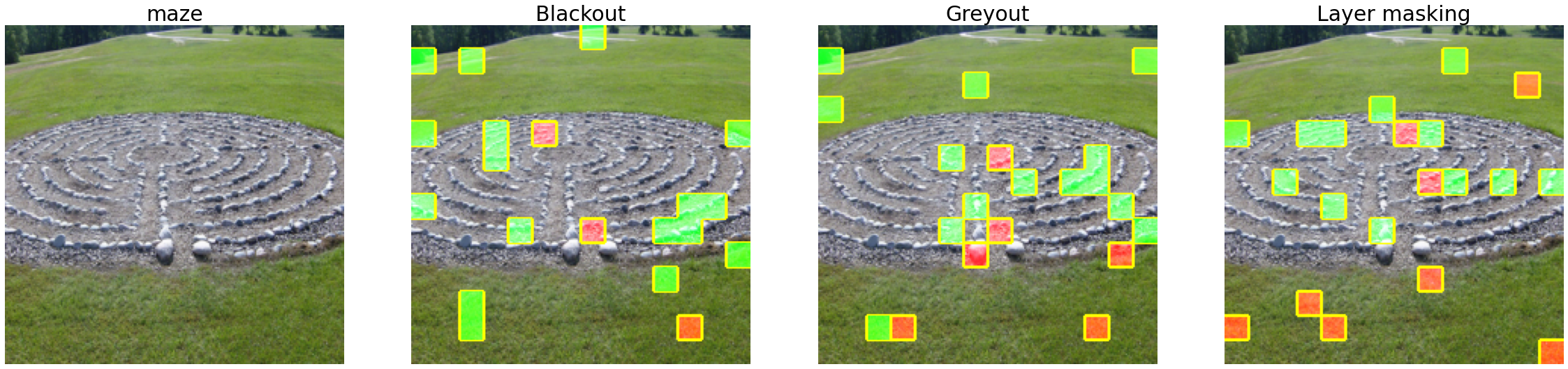}
    \includegraphics[width=\textwidth]{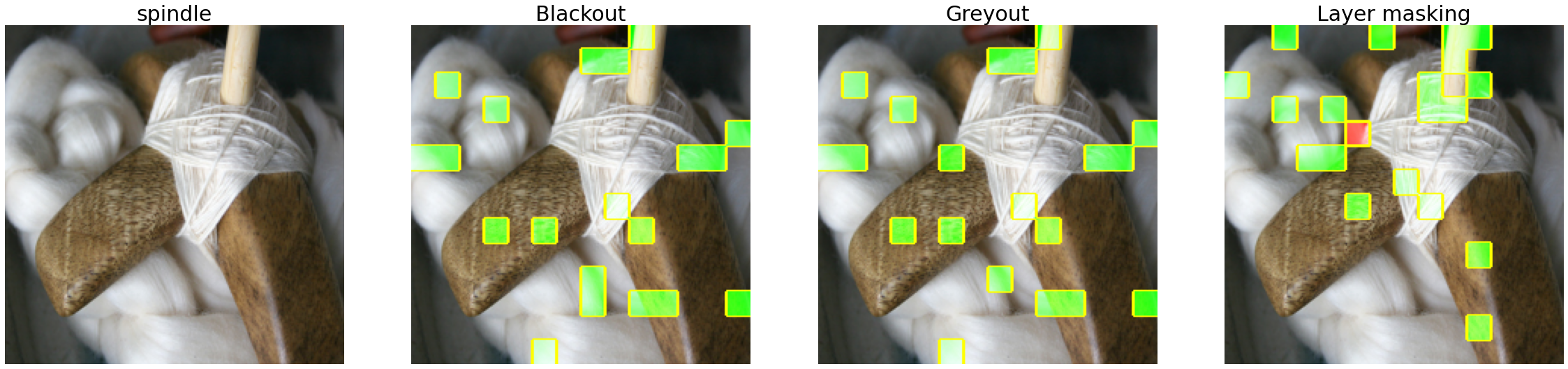}
    \includegraphics[width=\textwidth]{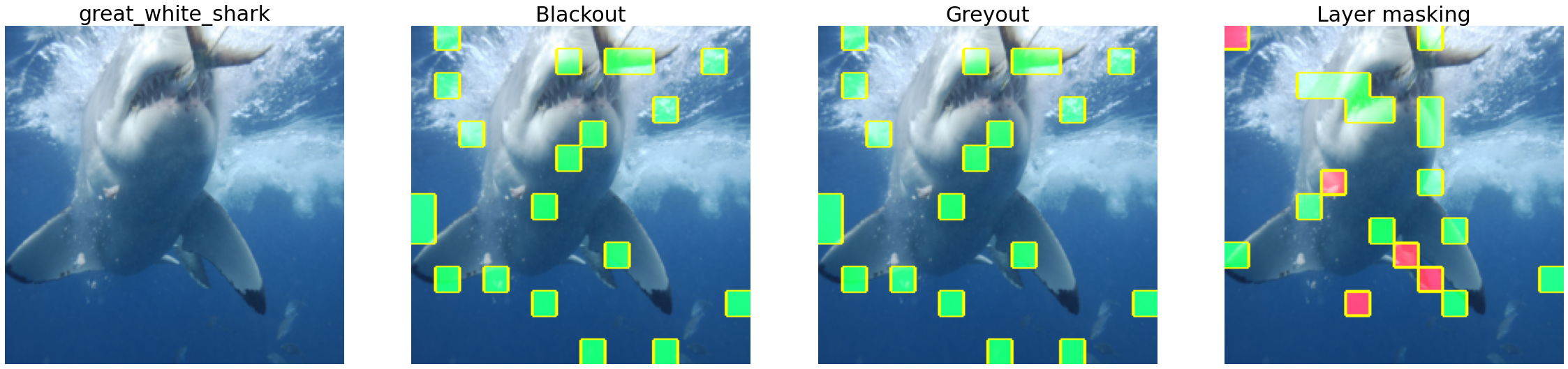}

    \caption{LIME scores using $16 \times 16$ segmentation (5 samples). Top 20 segments are highlighted}
    \label{fig:lime_16_ex_2}
\end{figure*}
\newpage

\newpage

\clearpage

\subsection{Quantitative}

We compute the same metrics (Top-20 ablation accuracy, Alignment score, Top-20 Jaccard similarity) for different architectures and segmentation algorithms. The metrics are  computed as follows:

\begin{enumerate}
    \item \textbf{Top-$k$ ablation accuracy}: As described in Strumfels et al, we choose the $k$ most important segments according to the explanation, remove them by substituting with a missingness approximation (we use grey), and compute the accuracy on the masked images. The more the accuracy drops, the better the explanations. Let $\vm'$ be a mask such that if pixel $(u, v)$ lies in the top $k$ features, then $\vm'[u, v] = 1$ otherwise $0$. Then, the top-$k$ ablation accuracy is the accuracy  when images are masked by $\vm'$ using a missingness approximation $t$ (we use grey): 
    $$\E_{(x, y) \sim D}[ \mathbf{1}[f(x \odot (1 - \vm') + \vm' \odot t) = y] ]$$
    \item \textbf{Alignment score}: Given a segmentation mask $\vm \in [0, 1]^{d \times d}$ for an image of dimension $d$, we derive the ``ground truth" $\vg$ for the explanation such that  $\vg_i = \sum_{(u, v) \in \text{patch } i} \left(\vm[u, v] - m_{avg} \right)$ where $m_{avg}$ is the mean of the segmentation mask. We can then measure how aligned the explanations are with the ground truth by computing the \emph{alignment score}, which is the cosine similarity between $g_i$ and $s_i$, or $$ \cos (\vg, \vs ) = \frac{\sum_i g_i s_i}{\sqrt{(\sum_i g_i^2)(\sum_i s_i^2)}}$$ The alignment score will be $1$ if the LIME explanation $\vs$ is perfectly aligned with $\vg$, and $-1$ if it is completely misaligned. 
    \item \textbf{Top-$k$ Jaccard similarity}: Take the top-$k$ most contributing features according to the explanation and compute a mask $\vm'$ such that if pixel $(u, v)$ lies in the top $k$ features, then $\vm'[u, v] = 1$ otherwise $0$. Then, we compute Jaccard similarity between the segmentation mask $\vm$ and $\vm'$ as $$\text{JaccSim}(\vm, \vm') =\frac{\sum_{u, v} \vm[u, v] \cdot \vm'[u, v]}{ \sum_{u, v}\mathbf{1}[\vm[u, v] + \vm'[u, v] > 0]}$$
\end{enumerate}

All of these metrics have their pros and cons. Top $k$ ablation accuracy does not require any supervision or ground truth, but has an undesirable dependence on the missingness approximation used to compute it. The alignment score is designed such that random attributions get a score of 0, but has an undesirable dependence on scale of the explanations. Top $k$ Jaccard similarity is not dependent on the scale, but only the relative ordering of importance of the features, but has a non-zero value for random features. Together, they give a more complete picture of the performance of LIME.

We report our results in \cref{tab:lime_metrics_full}. For Wide ResNet-50, AlexNet, SqueezeNet, and DenseNet, the performance of layer masking is the best across all metrics. For EfficientNet and MobileNet-v3, performance of layer masking is worse than greyout in top-$k$ ablation accuracy, but better in alignment score and top-$k$ Jaccard similarity.

\begin{table*}[ht!]
    \centering
    \begin{tabular}{@p{25mm}|^p{13mm}^p{10mm}^p{10mm}|^p{13mm}^p{10mm}^p{10mm}|^p{13mm}^p{10mm}^p{10mm}}
     \hline
                  & \multicolumn{3}{c|}{Top-20 ablation accuracy ($\downarrow$)} &  \multicolumn{3}{c|}{Alignment score  ($\uparrow$)} &  \multicolumn{3}{c}{Top-20 Jaccard similarity  ($\uparrow$)}  \\
                  &  Quickshift & $16 \times 16$ & SLIC & Quickshift & $16 \times 16$ & SLIC &  Quickshift & $16 \times 16$ & SLIC\\
                  \hline
                  \textbf{Wide ResNet-50} &&&&&&&&&\\
                Blackout      &  0.668 &  0.736 &  0.767 &  0.128 &  0.028 &  0.091 &  0.177 &  0.089 &  0.128 \\
Greyout       &  0.395 &  0.642 &  0.611 &  0.246 &  0.084 &  0.195 &  0.232 &  0.113 &  0.180 \\
Layer masking &  0.315 &  0.392 &  0.429 &  0.319 &  0.252 &  0.276 &  0.267 &  0.188 &  0.216 \\
\hline
    \textbf{AlexNet} &&&&&&&&&\\
       Blackout      &  0.550 &  0.506 &  0.681 &  0.039 &  0.006 &  0.020 &  0.139 &  0.085 &  0.097 \\
Greyout       &  0.375 &  0.488 &  0.531 &  0.114 &  0.014 &  0.074 &  0.189 &  0.089 &  0.124 \\
Layer masking &  0.181 &  0.256 &  0.331 &  0.209 &  0.200 &  0.187 &  0.240 &  0.167 &  0.188 \\
         \hline
         \textbf{SqueezeNet} &&&&&&&&&\\
         Blackout      &  0.479 &  0.552 &  0.615 &  0.058 &  0.002 &  0.031 &  0.154 &  0.081 &  0.101 \\
Greyout       &  0.307 &  0.547 &  0.568 &  0.124 &  0.015 &  0.075 &  0.195 &  0.087 &  0.129 \\
Layer masking &  0.224 &  0.234 &  0.281 &  0.197 &  0.194 &  0.186 &  0.235 &  0.167 &  0.189 \\
        \hline
        \textbf{DenseNet} &&&&&&&&&\\
        Blackout      &  0.562 &  0.745 &  0.682 &  0.156 &  0.029 &  0.122 &  0.203 &  0.089 &  0.149 \\
Greyout       &  0.276 &  0.589 &  0.495 &  0.273 &  0.099 &  0.234 &  0.259 &  0.122 &  0.196 \\
Layer masking &  0.312 &  0.359 &  0.500 &  0.301 &  0.261 &  0.290 &  0.277 &  0.195 &  0.220 \\
    \hline
    \textbf{MobileNet}&&&&&&&&&\\
    Blackout      &  0.562 &  0.896 &  0.719 &  0.214 &  0.072 &  0.173 &  0.225 &  0.108 &  0.168 \\
Greyout       &  0.365 &  0.526 &  0.536 &  0.237 &  0.167 &  0.207 &  0.231 &  0.159 &  0.182 \\
Layer masking &  0.547 &  0.656 &  0.599 &  0.258 &  0.203 &  0.241 &  0.249 &  0.168 &  0.201 \\
\hline
  \textbf{EfficientNet} &&&&&&&&&\\
Blackout      &  0.703 &  0.901 &  0.771 &  0.251 &  0.084 &  0.231 &  0.246 &  0.119 &  0.199 \\
Greyout       &  0.500 &  0.646 &  0.604 &  0.236 &  0.175 &  0.198 &  0.244 &  0.167 &  0.192 \\
Layer masking &  0.661 &  0.688 &  0.750 &  0.291 &  0.231 &  0.266 &  0.268 &  0.185 &  0.216 \\
\hline
    \end{tabular}
    \caption{Top-20 ablation accuracy, alignment score, and top-20 Jaccard similarity of LIME scores  over 200 random images }
    \label{tab:lime_metrics_full}
\end{table*}

\clearpage
\section{Other interesting properties of layer masking }

In this section, we identify some more properties of layer masking that are important for model interpretability.

\subsection{Linearity in masking:}
Consider a model equipped with a masking technique $f_m$ which acts on an input - mask pair $(\vx, \vm)$ and returns an output $\vy$ which depends only on the unmasked parts of the input. Then, we say that the model $f_m$ is linear in masking if 
 $f_m(\vx, \vm_1 + \vm_2) = f_m(\vx, \vm_1) +  f_m(\vx, \vm_2)$ for any two binary masks $\vm_1, \vm_2$ such that $\vm_1 \cdot \vm_2 = 0$.
This property is useful for interpretability methods like LIME which train a linear model on $(\vm, \vy)$ pairs and use its weights to explain the model prediction.
Modern vision models like CNNs and Vision Transformers are non-linear and include cross-interactions between features in $\vm_1$ and $\vm_2$. Thus, it is not possible to design a perfectly linear masking technique for these architectures, which means that only approximate linearity is possible. However, we can attempt to design more linear masking methods for each model architecture, and thus obtain more interpretable masking techniques. 

We measure linearity by sampling random images from ImageNet and dividing it into $N$ smaller square patches. We can then compute the cosine similarity between $f(\vx)$ and $\sum_{i=1}^N f_m(\vx, \vm_i)$ where $\vm_i$ corresponds to patch $i$ (\cref{tab:linearity_cos}). We find that layer masking is much more linear as compared to greying out or blacking out pixels, and in general, ResNet masking methods are more linear than corresponding methods for ViTs. Because the attention heads in ViTs introduce a lot of cross terms right from the beginning, including cross terms between distant patches, linearity in vision transformer masking is much lower than CNN masking. 

\begin{figure}[ht]
    \centering
    \includegraphics[width=0.6\textwidth]{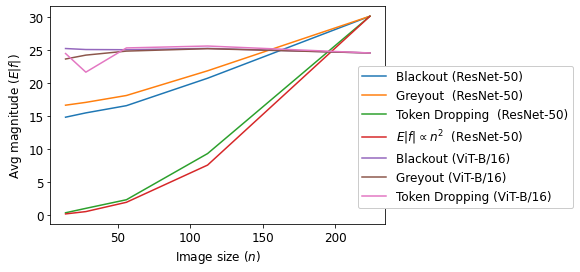}
   \caption{Mean magnitude of output feature vectors vs image size}
    \label{fig:mag_plot}
\end{figure}

We also find that in layer masking, $\E_\vx \|f_m(\vx, \vm)\|$ scales linearly with $|\vm|$. We test this by measuring the magnitude of $f_m(\vx, \vm)$ with $\vm$ as a mask for square patches of side length $n$, so that $\|\vm\| \propto n^2$. We observe in \cref{fig:mag_plot} that layer masking closely tracks the $n^2$ curve, which implies that  $\E_\vx \|f_m(\vx, \vm)\|$ scales almost linearly with $\|\vm\|$ for layer masking. However, the magnitude for ViT features remain approximately constant.

\begin{figure}[ht]
\centering
   \includegraphics[width=0.5\textwidth]{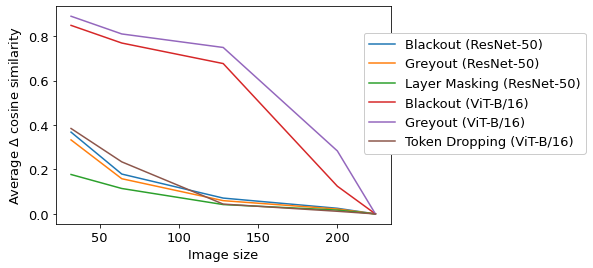}
  
   \caption{Average difference in cosine similarity  vs image size. Since model features of ViTs can be negative unlike ResNet-50, cosine similarity can vary from -1 to 1 }
   \label{fig:cos_sim_resize}
\end{figure}

\begin{table}[ht]
    \centering
    \begin{tabular}
{p{0.8cm}p{0.9cm}p{0.9cm}p{1.4cm}p{0.9cm}p{0.9cm}p{1.4cm}}
\toprule
 &  \multicolumn{3}{c}{ResNet-50 } & \multicolumn{3}{c}{ViT-B/16}\\
Patch size & Blackout & Greyout  & \textbf{Layer masking} & Blackout & Greyout  & Token dropping \\
\midrule

112 &   0.7975 &  0.8284 &       \textbf{ 0.9485} &   0.6707 &  0.7063 &        0.7043 \\
56  &   0.5124 &  0.5842 &        \textbf{0.8310} &   0.2202 &  0.2506 &        0.1929 \\
32  &   0.4282 &  0.4878 &        \textbf{0.7094} &   0.1377 &  0.1365 &        0.1426 \\
16  &   0.3848 &  0.4371 &        \textbf{0.6490} &   0.0912 &  0.0876 &        0.0877 \\
\bottomrule
\end{tabular}
\caption{Average cosine similarity between image features and their linear approximation}
   \label{tab:linearity_cos}

\end{table}

\newpage
\subsection{Avoidance of output collapse:}
 As the fraction of masked input approaches 1, it is desirable to avoid the model output collapsing to the same vector and thus not being sensitive enough to the unmasked features. To test for this, we take two random images $\vx_1$ and $\vx_2$ of size $224 \times 224$ and compute the cosine similarity between their model features, $c = \cos(f(\vx_1), f(\vx_2))$. Then, these images are resized to a smaller size $n$, and padded with zeros to recover the original size . We now have images $\vx_{1,n}$ and $\vx_{2,n}$ of size $224 \times 224$ and a mask of the same shape $\vm_n$ which is 1 for a region of size $n \times n$ and 0 elsewhere. We then measure the cosine similarity between $\vx_{1,n}$ and $\vx_{2,n}$ as $c_n = \cos(f_m(\vx_{1,n}, \vm_n), f_m(\vx_{2,n}, \vm_n))$ and plot $\E_{\vx_1, \vx_2}[c_n - c]$ as function of $n$ in \cref{fig:cos_sim_resize}.  We clearly see that as the image size is decreased, the cosine similarity changes much more for greyout or blackout as compared to layer masking for ResNet-50 or token dropping for ViTs.

\newpage

\clearpage
\section{Other baseline colors}

We also repeat the experiments in Section 4.1 with other baseline colors like red, blue and green. Grey baseline is included for reference. Segments are removed out in random order. We find that the best constant baseline is either greyout or average color of that image for both ResNet-50 and transformers. 

\begin{figure*}[ht!]
    
    \begin{subfigure}{0.45\textwidth}
         \includegraphics[width=\textwidth]{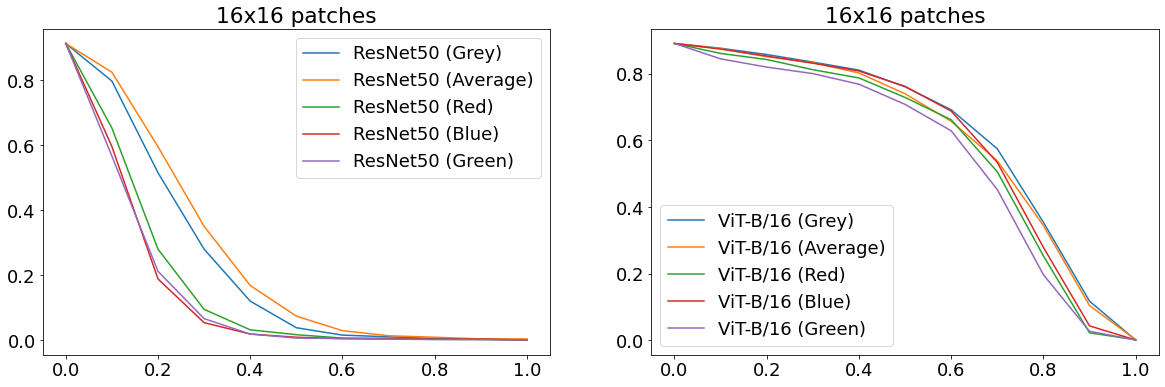}
        \includegraphics[width=\textwidth]{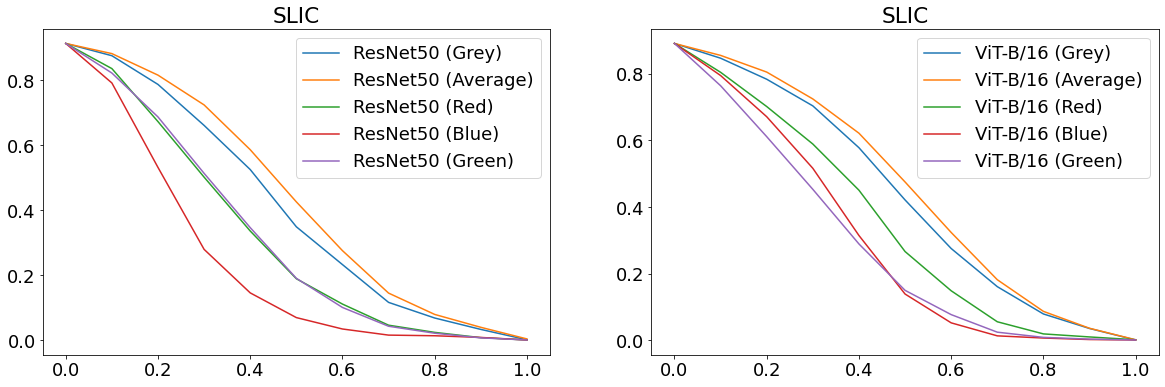}
         \caption{Accuracy}
         \label{fig:start_from_sal_slic_acc_plot_1}
     \end{subfigure}
     ~
     \begin{subfigure}[b]{0.45\textwidth}
         \includegraphics[width=\textwidth]{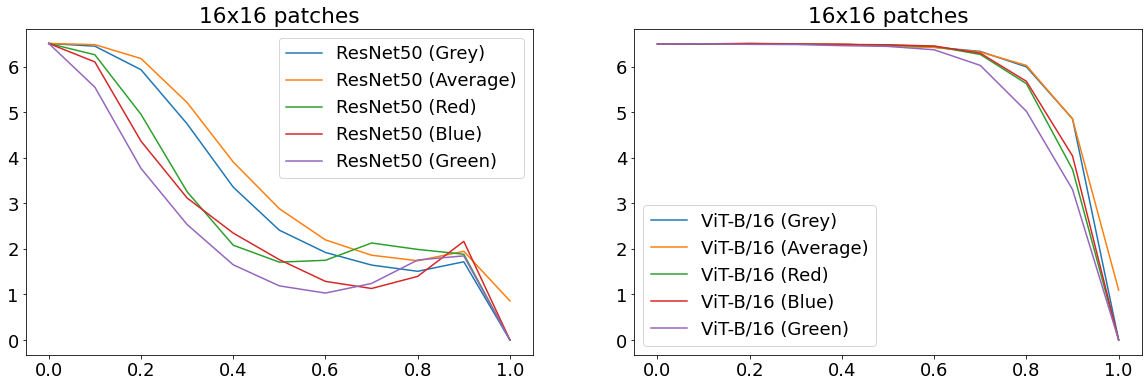}
         \includegraphics[width=\textwidth]{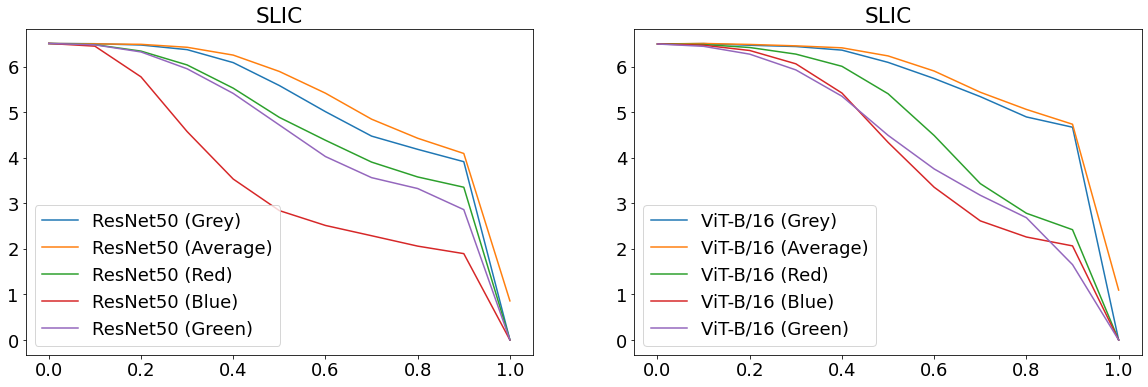}
         \caption{Class entropy}
         \label{fig:start_from_sal_quickshift_acc_plot_1}
     \end{subfigure}
    
     \begin{subfigure}{0.45\textwidth}
         \includegraphics[width=\textwidth]{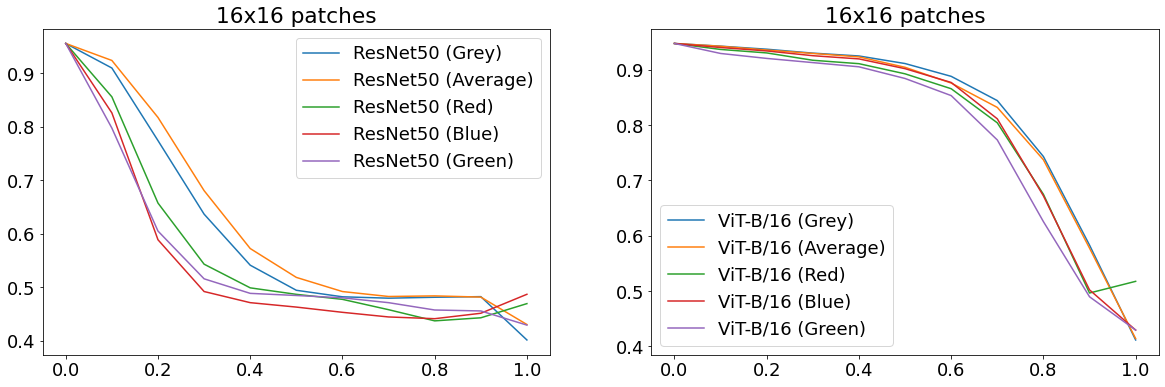}
        \includegraphics[width=\textwidth]{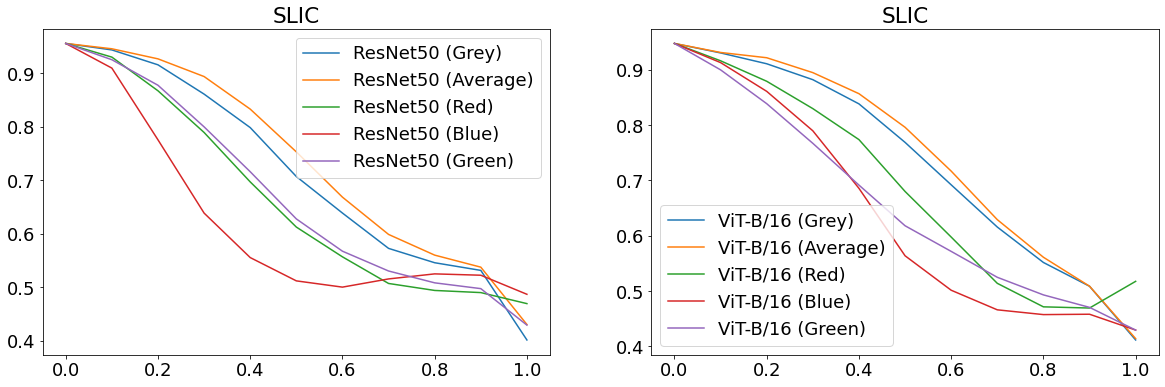}
         \caption{WordNet Similarity}
         \label{fig:start_from_sal_slic_acc_plot_2}
     \end{subfigure}
     ~
     \begin{subfigure}[b]{0.45\textwidth}
         \includegraphics[width=\textwidth]{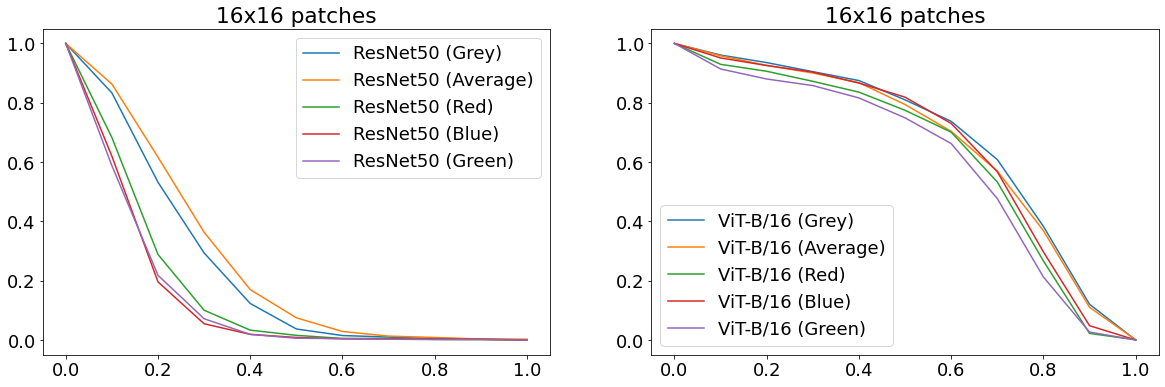}
         \includegraphics[width=\textwidth]{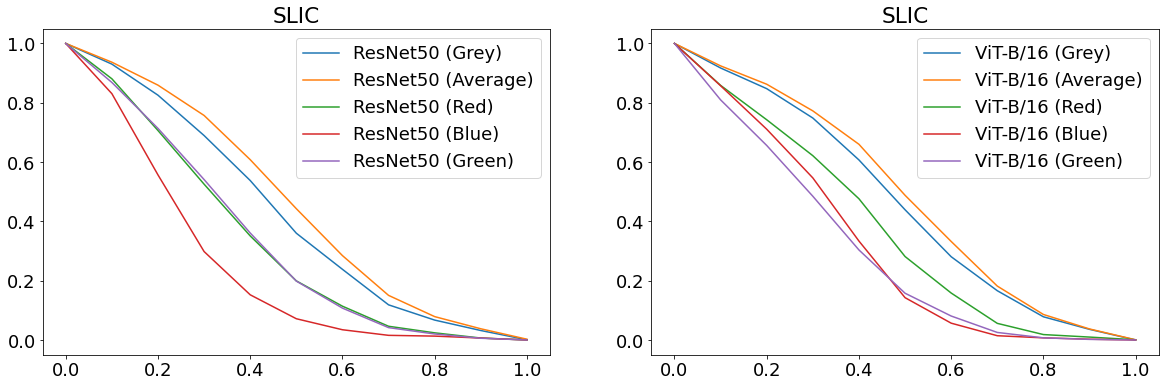}
         \caption{Unchanged predictions}
         \label{fig:start_from_sal_quickshift_acc_plot_2}
     \end{subfigure}
     
    \caption{Changes in model prediction for different model architectures}
    \label{fig:acc_degradation}
\end{figure*}

\end{document}